\documentclass{article} % For LaTeX2e
\usepackage{iclr2027_conference,times}

\usepackage{amsmath,amsfonts,bm}

\def\eqref#1{equation~\ref{#1}}
\def\1{\bm{1}}

\DeclareMathAlphabet{\mathsfit}{\encodingdefault}{\sfdefault}{m}{sl}
\SetMathAlphabet{\mathsfit}{bold}{\encodingdefault}{\sfdefault}{bx}{n}

\usepackage{hyperref}
\usepackage{url}
\usepackage[pdftex]{graphicx}

\usepackage{algorithm}
\usepackage{algorithmic}
\usepackage{amsmath}
\usepackage{amssymb}
\usepackage{mathtools}
\usepackage{amsthm}

\newtheorem{conjecture}{Conjecture}

\usepackage{booktabs}
\usepackage{multirow}

\definecolor{ColorComp}{HTML}{2196F3}
\definecolor{ColorMemo}{HTML}{FF9800}
\definecolor{ColorConf}{HTML}{F44336}
\definecolor{ColorGrok}{HTML}{4CAF50}

\definecolor{ForestGreenCustom}{RGB}{34, 139, 34}
\newcommand{\RevisionStart}{\begingroup\color{black}}
\newcommand{\RevisionEnd}{\endgroup}

\title{Misalignment of Low-Loss Regions \\ Causes Grokking}

\author{Yongding Tian\\
Computer Engineering Lab \\
Delft University of Technology\\
Delft, NL \\
\texttt{\{Y.Tian-3\}@tudelft.nl} \\
\And
Zaid Al-Ars \\
HDL TypeTech \\
Delft, NL \\
\AND
Maksim Kitsak \\
Network and Architecture Service \\
Delft University of Technology\\
Delft, NL \\
\AND
Peter Hofstee \\
IBM Infrastructure \\
TX, USA \\
}

\iclrfinalcopy % Show authors and suppress anonymous-review markings.
\begin{document}

\maketitle

\begin{abstract}
Grokking refers to the delayed emergence of validation-set generalization after a model has already overfit the training set. Although first observed in small algorithmic tasks trained with transformers, its underlying mechanism remains unsettled. In this work, we develop an analysis framework based on mode connectivity and the geometry of low-loss regions. The framework predicts that the standard modular-arithmetic setting does not always produce grokking: under a symmetry-preserving train/validation split, we observe a stable anti-grokking case in which validation performance does not recover. \RevisionStart This counterexample tests how several commonly discussed grokking indicators behave when grokking is absent.\RevisionEnd More broadly, our analysis framework and results further suggest that grokking arises when the low-loss regions induced by the training and validation partitions are misaligned. Once these regions become well aligned, training hyperparameters alone cannot produce grokking and the observed dynamics collapse to either trainable or non-trainable behavior.

\end{abstract}

\section{Introduction}

Grokking is the phenomenon in which a neural network generalizes to a validation set only after a long period of apparent overfitting to the training set. It was first reported by OpenAI when training a two-layer transformer on modular arithmetic tasks~\citep{power2022grokkinggeneralizationoverfittingsmall}. This behavior contrasts with the usual expectation that training and validation accuracy improve on comparable timescales, and it has been discussed as a form of emergent generalization in which a model eventually discovers the structure underlying the data~\citep{nanda2023progressmeasuresgrokkingmechanistic, wei2022emergentabilitieslargelanguage}.

Prior work has approached grokking from several perspectives, including Fourier circuit~\citep{nanda2023progressmeasuresgrokkingmechanistic}, representation learning~\citep{liu2022understandinggrokkingeffectivetheory}, numerical stability~\citep{prieto2025grokkingedgenumericalstability}, Hessian geometry~\citep{zhang2024a}, weight decay~\citep{xu2026grokgrokkingprovablegrokking}, the Slingshot effect~\citep{thilak2022the}, and information leakage~\citep{carvalho2025grokkingexplainedstatisticalphenomenon}. These studies reveal useful patterns related to grokking, but they do not provide a full causal account of why grokking occurs.

In this paper, we develop a geometric framework for analyzing grokking, based on empirical studies of Mode Connectivity (MC)~\citep{pmlr-v80-draxler18a, tian2025connectingindependentlytrainedmodes}. Singular Learning Theory (SLT) suggests that Low-Loss Regions (LLRs)\footnote{The sets of model weights whose loss is below a specified threshold.} in neural-network parameter spaces are better understood as extended valleys than as isolated basins~\citep{Visualizing_the_Loss_Landscape_of_Neural_Nets, ainsworth2023git,lion2024goodsinglebasin}. MC provides empirical tools for identifying low-loss paths between independently trained minima.

Our analysis framework predicts that the OpenAI modular-arithmetic setting does not necessarily produce grokking. We validate this prediction by constructing an anti-grokking split that reliably prevents delayed validation improvement while leaving the model architecture, optimizer, dataset size, and task family unchanged. \RevisionStart We then use this case to examine whether commonly discussed indicators also occur when grokking is absent. Several indicators do not by themselves predict this absence, suggesting that they should be interpreted as correlates rather than complete causal explanations. \RevisionEnd

Our analysis suggests that grokking requires two ingredients. First, the low-loss regions defined by the training and validation partitions must be sufficiently misaligned. Second, the training trajectory must enter the overlap of these regions in a delayed manner, a condition shaped by hyperparameters such as learning rate and weight decay. We regard the first ingredient as the root cause because grokking does not occur when these low-loss regions are well aligned, whereas the second ingredient controls how training moves through the low-loss regions. In practical terms, low-loss regions are easier to misalign when the information carried by the training partition is small relative to the expressive capacity of the model.

This paper makes the following contributions:
\begin{enumerate}
    \item We provide a geometric analysis framework that accounts for several proposed causes of grokking through the misalignment of low-loss regions.
    \item \RevisionStart Based on this framework, we identify counterexamples within the OpenAI grokking setup that do not exhibit grokking. These anti-grokking cases show that several commonly used indicators are not sufficient on their own to predict grokking.\RevisionEnd
    \item Based on this framework, we show that low-loss regions shrink and become more aligned as the information carried by the training partition increases. When the low-loss regions are fully aligned, grokking cannot occur.
\end{enumerate}

The paper is organized as follows. Section~\ref{sec:terminology_existing_work} introduces the terminology and reviews prior work on grokking and mode connectivity. Section~\ref{sec:framework_and_anti_grokking} presents our low-loss-region framework and the anti-grokking example, showing that grokking does not always occur in the OpenAI grokking setup. Section~\ref{sec:alignment_of_low_loss_region} studies the alignment of low-loss regions. Section~\ref{sec:discussion} discusses the implications of these results, and Section~\ref{sec:conclusion} concludes the paper.

\section{Terminology and Existing Work}
\label{sec:terminology_existing_work}

This section introduces the terminology used throughout the paper, reviews the OpenAI modular-arithmetic setup that we reuse in our experiments, summarizes several influential grokking explanations, and then describes mode connectivity, which is the empirical starting point for our framework.

\subsection{Common Terminology}
Consider a model architecture $\mathcal{M}$ with $d$ trainable parameters, a fixed dataset $\mathcal{D}$, and a loss function $\mathcal{L}$. Each parameter vector $P \in \mathbb{R}^d$ induces a loss value on $\mathcal{D}$. For a threshold $L_{\text{th}}$, we define the corresponding low-loss region $S_{L \leq L_{\text{th}}}$ as
\begin{align} \label{eq:define_low_loss_region}
    S_{L \leq L_{\text{th}}} \coloneqq \{ P \in \mathbb{R}^d \mid \mathcal{L}(P, \mathcal{D}) \leq L_{\text{th}} \} \hspace{1em}
\end{align}
This region is determined by the model architecture $\mathcal{M}$, dataset $\mathcal{D}$, loss function $\mathcal{L}$, and threshold $L_{\text{th}}$. It is independent from the optimizer and training hyperparameters. In principle, evaluating the loss at every point in parameter space would determine the boundary of this region. We therefore write Equation~\ref{eq:define_low_loss_region} as the following mapping:
\begin{align} \label{eq:define_LLR_from_model_architecture_dataset}
    S_{L \leq L_{\text{th}}} = \text{LLR}(\mathcal{M}, \mathcal{D}, \mathcal{L}, L_{\text{th}})
\end{align}
where LLR stands for Low-Loss Region. Since the loss function is coupled with the architecture and the threshold is adjusted to select a contour level, we use the concise notation $S_{L \leq L_{\text{th}}} = \text{LLR}(\mathcal{M}, \mathcal{D})$ when these choices are clear from context.

We use the following notation for the overlap of the low-loss regions induced by two datasets\footnote{The threshold $L_{\text{th}}$ is omitted for readability.}:
\begin{align*}
    \text{LLR}(\mathcal{M}, \mathcal{D}_1 + \mathcal{D}_2) & \coloneqq
    \{ 
    P \in \mathbb{R}^d \mid 
    P \in \text{LLR}(\mathcal{M},\mathcal{D}_1) 
    \land 
    P \in \text{LLR}(\mathcal{M},\mathcal{D}_2) 
    \}
\end{align*}
Similarly, we define the part of $\text{LLR}(\mathcal{M}, \mathcal{D}_1)$ that does not overlap with $\text{LLR}(\mathcal{M}, \mathcal{D}_2)$ as
\begin{align*}
    \text{LLR}(\mathcal{M}, \mathcal{D}_1 + !\mathcal{D}_2) & \coloneqq
    \{ 
    P \in \mathbb{R}^d \mid 
    P \in \text{LLR}(\mathcal{M},\mathcal{D}_1) 
    \land 
    P \notin \text{LLR}(\mathcal{M},\mathcal{D}_2) 
    \}
\end{align*}

\subsection{Set Up Grokking Experiments}

Grokking was first observed when training a simple transformer on a modular arithmetic dataset~\citep{power2022grokkinggeneralizationoverfittingsmall}. We reuse this experimental setup and denote the corresponding model architecture by $\mathcal{M}_{grok}$. The full architecture and training recipe are provided in Appendix~\ref{appendix:model_arch_and_train_recipe}.

\begin{figure}[htbp]
    \centering
    \includegraphics[width=0.9\linewidth]{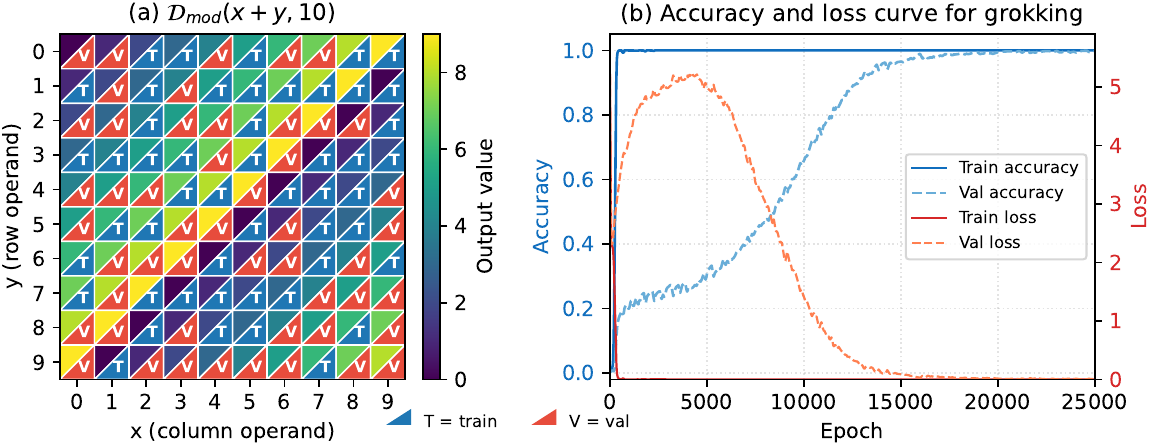}
    \caption{Two-panel illustration of the grokking setup. (a) Illustration of the modular dataset $\mathcal{D}_{mod}(x+y, 10)$. The x-axis and y-axis represent the values of $x$ and $y$ for the mapping function $f(x,y)=x+y$. Each square indexed by $(x,y)$ contains two triangles: the upper-left triangle gives the value of $f(x,y)$, and the lower-right triangle indicates whether the sample belongs to the training or validation partition. This small example is shown only for illustration purposes, and our experiments use the larger dataset $\mathcal{D}_{mod}(x+y, 97)$. (b) Illustration of grokking. Validation accuracy increases only long after the model has overfit the training partition. The result is reproduced with $\mathcal{M}_{grok}$ on $\mathcal{D}_{mod}(x+y, 97)$.}
    \label{fig:grokking_setup}
\end{figure}

The modular arithmetic dataset maps integer pairs $(x,y)$ to $f(x,y) \bmod m$, where $m$ is the modulus and $x,y \in \{0,\ldots,m-1\}$. We write this dataset as $\mathcal{D}_{mod}(f(x,y), m)$. Modular addition is the most commonly studied function in grokking research~\citep{nanda2023progressmeasuresgrokkingmechanistic, thilak2022the}. We illustrate $\mathcal{D}_{mod}(x+y, 10)$ in Figure~\ref{fig:grokking_setup}(a).

The original OpenAI setup uses $\mathcal{D}_{mod}(x+y, 97)$ with randomly $50\%$ of the samples assigned to training and the remainder assigned to validation~\citep{power2022grokkinggeneralizationoverfittingsmall}. We follow this setup and reproduce grokking\footnote{The full training recipe is provided in Appendix~\ref{appendix:model_arch_and_train_recipe}.}, see Figure~\ref{fig:grokking_setup}(b).

\subsection{Existing Work on Grokking}

Prior work has proposed several mechanisms and empirical indicators associated with grokking. We summarize the main indicator claims most relevant to our analysis below.

\begin{figure}[htbp]
    \centering
    \includegraphics[width=0.99\linewidth]{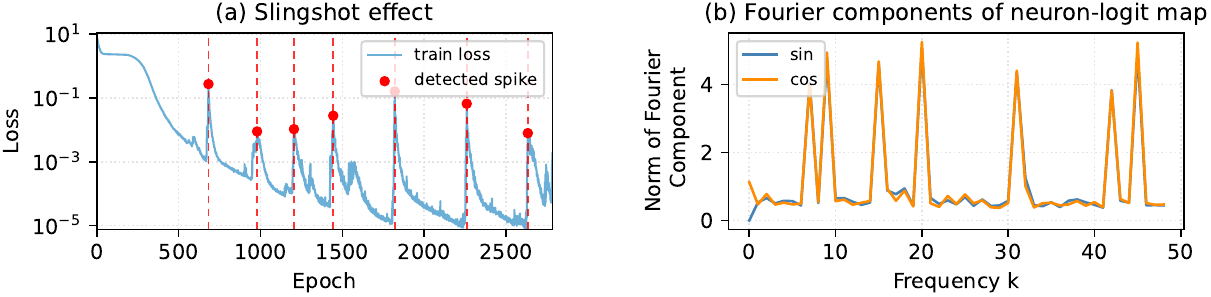}
    \caption{Two-panel illustration of grokking indicators. (a) Slingshot effect: red dots mark spike locations on the training-loss curve. The curve is taken from the experiment in Figure~\ref{fig:grokking_setup}(b) and plotted on a log scale to make the spikes visible. Prior work reports that such spikes are positively correlated with grokking~\citep{thilak2022the}. (b) Fourier components of the neuron-logit map for a well-trained grokked model. The model is obtained at the end of the experiment in Figure~\ref{fig:grokking_setup}(b). Prior work suggests that the formation of this Fourier circuit can be used as a progress measure for grokking~\citep{nanda2023progressmeasuresgrokkingmechanistic}.}
    \label{fig:grokking_indicators}
\end{figure}

\begin{enumerate}
    \item \label{existing:slinghots}
    \RevisionStart
    \textbf{Slingshot effect}
    Thilak et al.~\citep{thilak2022the} report that grokking almost exclusively occurs near the onset of Slingshots and is absent in runs without them. We treat this as an empirical association between Slingshots and grokking.
    \RevisionEnd
    
    \item \label{existing:fourier} \textbf{Fourier circuit} In transformers trained on $\mathcal{D}_{mod}(x+y, m)$, grokking has been linked to the formation of a Fourier-based generalizing circuit~\citep{nanda2023progressmeasuresgrokkingmechanistic}. This circuit has also been proposed as a progress measure for grokking. Figure~\ref{fig:grokking_indicators}(b) visualizes the corresponding Fourier components.
    \item \label{existing:training_set_size} \textbf{Training partition threshold} Several studies argue that the training partition must exceed a critical size before a model can learn a generalizing representation and exhibit grokking~\citep{varma2023explaininggrokkingcircuitefficiency,liu2022understandinggrokkingeffectivetheory,zhu2024criticaldatasizelanguage,pmlr-v267-abramov25a}.
    \item \label{existing:grok_in_other_dataset_arch} \textbf{Grokking in MNIST/CIFAR} Grokking is not limited to transformers or modular arithmetic. It has also been observed when training Vision Transformers (ViTs) on CIFAR and MLPs on MNIST~\citep{liu2022understandinggrokkingeffectivetheory,6296535,cifar10_dataset,thilak2022the,dosovitskiy2021an}.
\end{enumerate}

\subsection{Existing Work on Mode Connectivity}

Mode connectivity is the empirical observation that two independently trained models can be connected by a low-loss path~\citep{pmlr-v80-draxler18a}. This property has been observed for ResNet~\citep{he2015deep}, VGG~\citep{simonyan2015deepconvolutionalnetworkslargescale}, and DenseNet~\citep{huang2018denselyconnectedconvolutionalnetworks} models trained on CIFAR-10. More recently, the Low-Loss Path Finding (LLPF) method extended this observation to transformer architectures~\citep{tian2025connectingindependentlytrainedmodes}.

We empirically validate mode connectivity for $\mathcal{M}_{grok}$ on $\mathcal{D}_{mod}(x+y,97)$ using LLPF. Configuration details and results are provided in Appendix~\ref{appendix:empirical_validation_of_mode_connectivity}. For two independently trained $\mathcal{M}_{grok}$ models, the maximum loss along the discovered path remains below $1.5\times10^{-3}$ and the minimum accuracy remains above $99.85\%$. Repeating this experiment ten times yields only small standard deviation across runs. These empirical results motivate Conjecture~\ref{conj:low_loss_region_is_path_connected}.
\begin{conjecture}
    \label{conj:low_loss_region_is_path_connected}
    The low-loss region $\text{LLR}(\mathcal{M}_{grok}, \mathcal{D}_{mod}(x+y,97),\mathcal{L},1.5\times10^{-3})$ is a single path-connected region.
\end{conjecture}
This conjecture is consistent with the LLPF interpretation that repeated success across different pairs of modes trained on random seeds provides evidence for a path-connected low-loss region. The ``path'' should be understood as evidence of a higher-dimensional connected manifold rather than a literal one-dimensional structure: LLPF identifies one-dimensional curves, but the underlying connected set is expected to have dimension greater than one~\citep{pmlr-v139-benton21a}.

There is also theoretical support for this perspective, please see Appendix~\ref{appendix:slt}. We also clarify that MC should not be conflated with linear mode connectivity, see Appendix~\ref{appendix:mc_vs_lmc}.

\section{Analysis Framework and Anti-Grokking Example}
\label{sec:framework_and_anti_grokking}

This section introduces our analysis framework and uses it to identify a corner case in the OpenAI modular-arithmetic setup where grokking does not occur. We call this behavior \emph{anti-grokking}. We then use the anti-grokking case to test whether several existing grokking indicators can predict the absence of grokking.

\subsection{Analysis Framework}
Consider a dataset $\mathcal{D}$ split into a training partition $\mathcal{D}_{train}$ and a validation partition $\mathcal{D}_{val}$. These two partitions induce two low-loss regions, $\text{LLR}(\mathcal{M},\mathcal{D}_{train})$ and $\text{LLR}(\mathcal{M},\mathcal{D}_{val})$. Because $\mathcal{D}_{train} \neq \mathcal{D}_{val}$, the corresponding low-loss regions are not completely identical.

In the grokking experiment shown in Figure~\ref{fig:grokking_setup}(b), the training partition is randomly sampled from $\mathcal{D}_{mod}(x+y,97)$ and contains $50\%$ of the whole dataset. We therefore expect the two induced low-loss regions to have a comparable volume:
$V(\text{LLR}(\mathcal{M},\mathcal{D}_{train})) \approx V(\text{LLR}(\mathcal{M},\mathcal{D}_{val}))$.

When the two regions are comparable in volume but not identical, some weight vectors $P$ may lie in $\text{LLR}(\mathcal{M},\mathcal{D}_{train})$ without lying in $\text{LLR}(\mathcal{M},\mathcal{D}_{val})$. We call this set the \emph{non-overlap low-loss region} and write the non-empty condition as
\begin{align}
    \label{eq:partial_llr_is_not_empty}
    \text{LLR}(\mathcal{M}, \mathcal{D}_{train} + !\mathcal{D}_{val} )  \neq \varnothing .
\end{align}
The same reasoning applies with $\mathcal{D}_{train}$ and $\mathcal{D}_{val}$ interchanged.

The second question is whether the overlap between the two low-loss regions is non-empty:
\begin{align}
    \label{eq:overlap_llr_is_not_empty}
    \text{LLR}(\mathcal{M}, \mathcal{D}_{train} + \mathcal{D}_{val} ) \neq? \varnothing .
\end{align}
For $\mathcal{M}_{grok}$ and $\mathcal{D}_{mod}(x+y,97)$, this overlap is not empty because Figure~\ref{fig:grokking_setup}(b) shows that some parameter vectors achieve low loss on both training and validation partitions. In other settings, however, the overlap may be empty, for example when the architecture is inappropriate for the task or when the labels are randomly generated.

\begin{figure}[htbp]
    \centering
    \includegraphics[width=0.4\linewidth]{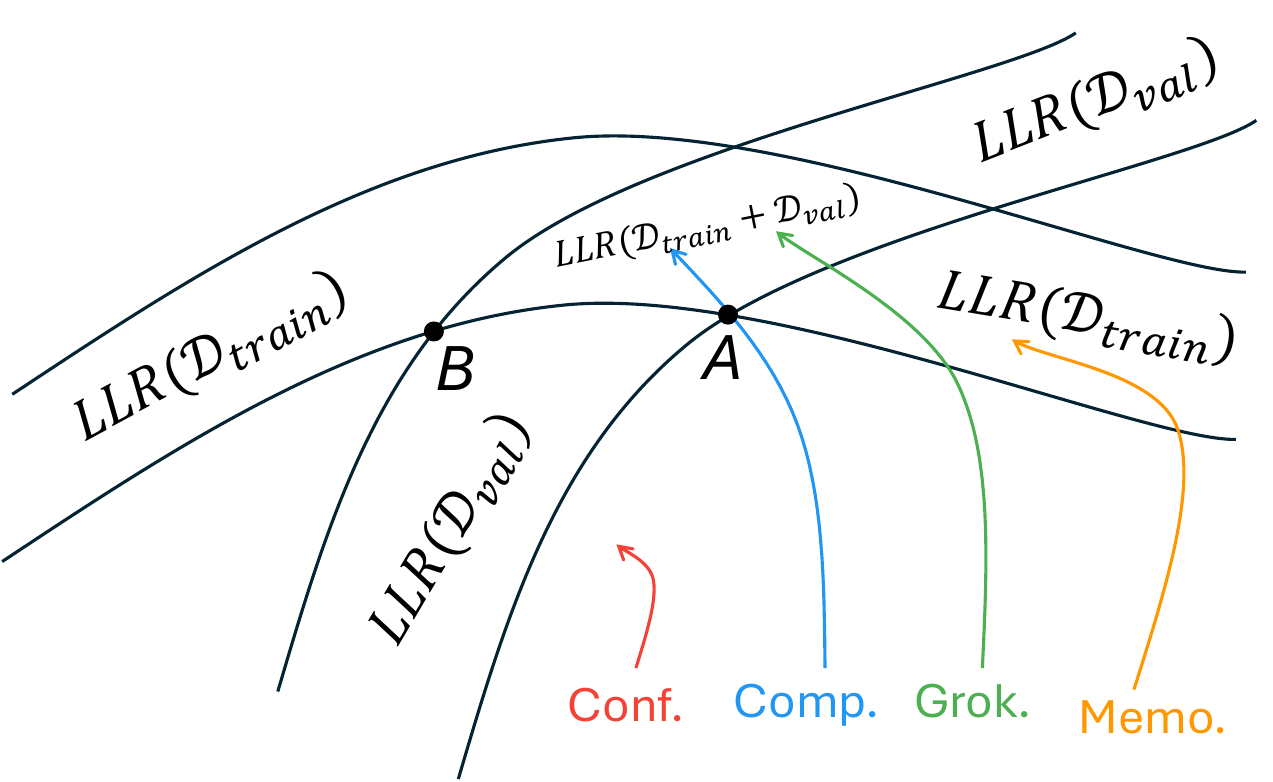}
    \caption{Geometric illustration of $\text{LLR}(\mathcal{M}_{grok}, \mathcal{D}_{mod}(x+y, 97))$ in the weight space. Motivated by Conjecture~\ref{conj:low_loss_region_is_path_connected}, we draw connected paths representing $\text{LLR}(\mathcal{D}_{train})$ and $\text{LLR}(\mathcal{D}_{val})$. Equations~\ref{eq:partial_llr_is_not_empty} and~\ref{eq:overlap_llr_is_not_empty} indicate that both overlap and non-overlap regions exist for the grokking setup, so the two low-loss regions are not aligned. The four learning trajectories (\textcolor{ColorComp}{Comprehension}, \textcolor{ColorGrok}{Grokking}, \textcolor{ColorMemo}{Memorization}, and \textcolor{ColorConf}{Confusion}) correspond to the learning phases identified in~\citep{liu2022understandinggrokkingeffectivetheory}.}
    \label{fig:grokking_explaination}
\end{figure}

Combining Conjecture~\ref{conj:low_loss_region_is_path_connected} with Equations~\ref{eq:partial_llr_is_not_empty} and~\ref{eq:overlap_llr_is_not_empty} yields the geometric illustration in Figure~\ref{fig:grokking_explaination}. This illustration is schematic and does not encode dimensionality~\footnote{For example, point $A$ should not be interpreted as a zero-dimensional point in $\mathbb{R}^d$, and curve $\overset{\frown}{AB}$ should not be interpreted as a literal one-dimensional curve. Likewise, the figure does not imply that exactly one $\text{LLR}(\mathcal{D}_{train})$ path intersects exactly one $\text{LLR}(\mathcal{D}_{val})$ path. Several high-dimensional branches may intersect in the overlap region.}.

Existing grokking work identifies four learning phases~\citep{liu2022understandinggrokkingeffectivetheory}:
\begin{itemize}
    \item \textcolor{ColorComp}{Comprehension}: accuracy on $\mathcal{D}_{train}$ and $\mathcal{D}_{val}$ increases at approximately the same time.
    \item \textcolor{ColorGrok}{Grokking}: validation accuracy increases much later than training accuracy.
    \item \textcolor{ColorMemo}{Memorization}: only training accuracy increases.
    \item \textcolor{ColorConf}{Confusion}: neither training nor validation accuracy increases.
\end{itemize}
These trajectories are also shown in Figure~\ref{fig:grokking_explaination}. Because the illustration is dimension-free, it should not be read as implying that \textcolor{ColorComp}{comprehension} is less likely than other phases merely because the drawn trajectory appears to pass through a point.

The grokking trajectory in Figure~\ref{fig:grokking_explaination} has two stages: (1) entering $\text{LLR}(\mathcal{D}_{train})$ and (2) moving from $\text{LLR}(\mathcal{D}_{train})$ into $\text{LLR}(\mathcal{D}_{train}+\mathcal{D}_{val})$. 
\RevisionStart
The first stage is the ordinary outcome of minimizing training loss, whereas the second requires explaining why optimization continues after training accuracy has saturated and why that continued motion can reach the overlap region. Appendix~\ref{appendix:cause_of_moving_to_overlap} gives a loss-based explanation: correctly classified training examples can still have nonzero cross-entropy, so the training loss can keep decreasing after training accuracy reaches $100\%$. This continued optimization can move the parameters through the training-only region and into the overlap when that region is accessible. Since low-loss regions are determined by $\mathcal{M}$ and $\mathcal{D}$, modifying the train/validation split can therefore alter whether this transition is possible.
\RevisionEnd

\subsection{Anti-Grokking Example}
Based on our analysis framework and prior discussion, we find one way to prevent the second stage using the notion of surprisal from Shannon information theory~\citep{shannon1948mathematical}. Each sample in a binary modular arithmetic dataset maps two variables $a$ and $b$ to an output $c$. For $\mathcal{D}_{mod}(x+y,97)$, the function $f(x,y)=x+y$ satisfies the symmetry $f(a,b)=f(b,a)$. We treat this symmetry as task \emph{knowledge}. If the fact $f(a,b)=c$ is known together with this symmetry, then observing $f(b,a)=c$ should have lower surprisal than it would without the symmetry. This suggests that a train/validation split can preserve the same training ratio while reducing the effective surprisal of the training partition. Empirically, we find such a corner case: grokking is eliminated by enforcing this symmetry in the partition. We call this the \emph{anti-grokking} example.

\begin{figure}[htbp]
    \centering
    \includegraphics[width=0.8\linewidth]{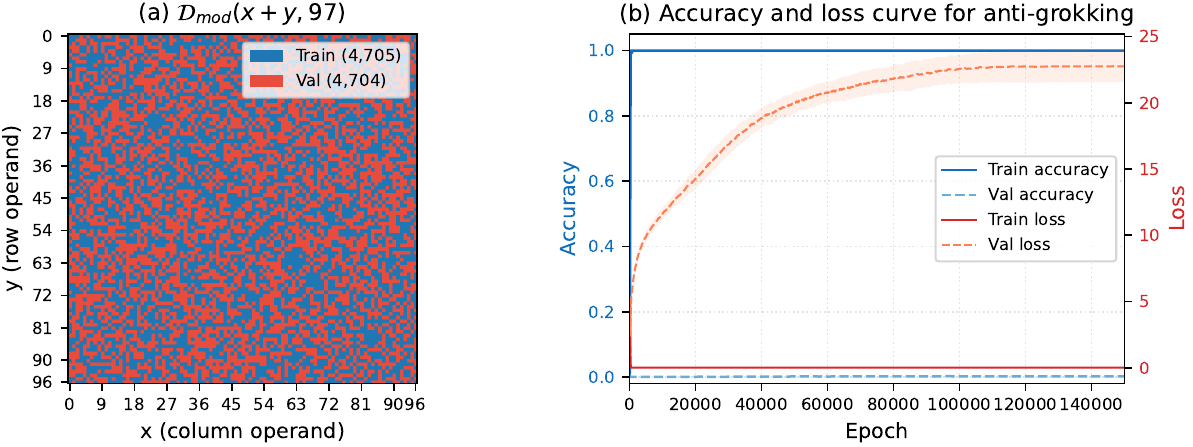}
    \caption{Anti-grokking setup. (a) Dataset partition for the anti-grokking example on $\mathcal{D}_{mod}(x+y,97)$. Unlike a random partition, samples located at $(a,b)$ and $(b,a)$ always belong to the same partition. The numbers $4705$ and $4704$ indicate the sample counts in the two partitions. (b) Training $\mathcal{M}_{grok}$ with the same hyperparameters on the split shown in Figure~\ref{fig:anti_grokking_setup}(a) does not produce grokking: validation accuracy remains low, and validation loss does not decrease. We repeat this experiment with ten random seeds and different splits. The shaded region indicates the standard deviation across runs.}
    \label{fig:anti_grokking_setup}
\end{figure}

We randomly generate a train/validation partition for $\mathcal{D}_{mod}(x+y,97)$ such that paired samples $f(a,b)$ and $f(b,a)$ always belong to the same partition, as shown in Figure~\ref{fig:anti_grokking_setup}(a). Training $\mathcal{M}_{grok}$ with the same hyperparameters on this anti-grokking split does not produce grokking, as shown in Figure~\ref{fig:anti_grokking_setup}(b). The only change relative to Figure~\ref{fig:grokking_setup}(b) is the train/validation split.

Section~\ref{sec:terminology_existing_work} summarized three indicators often associated with grokking: (1) the Slingshot effect, (2) the Fourier-based generalization circuit, and (3) a critical training-set size. 
\RevisionStart
We revisit these indicators in the anti-grokking example to examine whether they are also sufficient to guarantee grokking.
\RevisionEnd

\begin{figure}[htbp]
    \centering
    \includegraphics[width=0.99\linewidth]{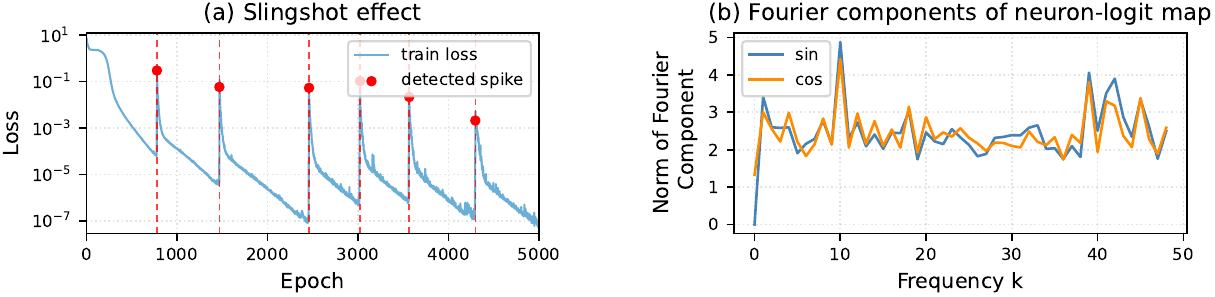}
    \caption{Anti-grokking diagnostics. (a) A Slingshot pattern similar to Figure~\ref{fig:grokking_indicators}(a) is observed in the anti-grokking example. Thus, the Slingshot effect is not by itself a faithful explanation of grokking. (b) Compared with Figure~\ref{fig:grokking_indicators}(b), several frequencies still have larger components than others, but the concentration is much weaker than in the grokking case.}
    \label{fig:anti_grokking_indicators}
\end{figure}

\RevisionStart
In the anti-grokking example, Slingshot spikes are still observed, see Figure~\ref{fig:anti_grokking_indicators}(a). Thilak et al.'s Figure 5 already shows that Slingshot effects and grokking can separate across datasets and training configurations, including cases with Slingshot effects but no grokking~\citep{thilak2022the}. Our experiment provides a complementary observation: this relationship can also change within the same task, model, optimizer, and complete dataset when only the train/validation split is changed. Thus, our experiment shows that a Slingshot is not sufficient by itself to guarantee grokking.\RevisionEnd The Fourier component plot in Figure~\ref{fig:anti_grokking_indicators}(b) also shows that some frequencies have larger components than others, although the concentration is much weaker than in Figure~\ref{fig:grokking_indicators}(b). However, using this signal as a progress measure is misleading in this case: Figure~\ref{fig:anti_grokking_setup}(b) shows no progress toward grokking, whereas the Fourier components suggest slight progress.

The claim that the training set must exceed a size threshold is also incomplete, because the anti-grokking example keeps the training-set size fixed. A more precise statement is that \emph{grokking requires the training partition to carry sufficient effective information to recover the task}. This effective information entropy is not determined solely by the raw number of samples: it should account for task knowledge that makes some samples informative about others. Appendix~\ref{appendix:effective_information_entropy} operationalizes this idea for the symmetry $f(a,b)=f(b,a)$ and shows that symmetry-aware training coverage provides a useful empirical proxy for effective information entropy. Prior work has used entropy-based notions for datasets~\citep{NEURIPS2025_39d3b980, Schwalbe-Koda2025, zhang2020dime}, but has not explicitly incorporated task knowledge in this way. A formal entropy measure that incorporates task knowledge is left for future work.

Finally, the requirement that samples at $(a,b)$ and $(b,a)$ appear in the same partition is not an all-or-nothing condition. In the anti-grokking split, $100\%$ of such pairs share a partition. As this ratio is reduced toward $70\%$, grokking gradually reappears, see Appendix~\ref{appendix:anti_grokking_example_with_other_ratios}. Anti-grokking is rarely observed under the original random partitioning policy because the same-partition ratio concentrates around $50\%$.

\section{Alignment of Low-Loss Regions}
\label{sec:alignment_of_low_loss_region}

This section analyzes how low-loss regions change as the dataset becomes larger. We first argue that adding data imposes additional constraints and therefore tends to shrink low-loss regions. We then empirically show that, as these regions shrink, the train and validation low-loss regions become increasingly aligned. Under such alignment, grokking disappears and training dynamics collapse to either trainable or non-trainable.

\subsection{Shrinking of Low-Loss Region}
The volume of a low-loss region should decrease, or at least not increase, as the dataset grows. When one dataset is a subset of another, this monotonic relation can be written as
\begin{align}
\label{eq:shrinking_of_low_loss_region}
    \text{LLR}(\mathcal{M}, \mathcal{D}_1) \subseteq \text{LLR}(\mathcal{M}, \mathcal{D}_2) \quad \text{if } \mathcal{D}_2 \subseteq \mathcal{D}_1
\end{align}
Equation~\ref{eq:shrinking_of_low_loss_region} follows from interpreting each data point as a constraint on the parameter vector. If $\mathcal{D}_1$ contains every sample in $\mathcal{D}_2$ plus additional samples, then any parameter vector that has low loss on $\mathcal{D}_1$ also satisfies the constraints imposed by $\mathcal{D}_2$. This intuition is consistent with empirical work showing that models typically incur higher loss on additional out-of-distribution data than on in-distribution data~\citep{zhang2017understandingdeeplearningrequires, salman2019overfittingmechanismavoidancedeep}. However, with average loss, this relationship should be understood as a geometric heuristic rather than a strict theorem, because errors on different samples can compensate for one another.

We slightly modify the modular-dataset construction so that it satisfies the nested-dataset condition in Equation~\ref{eq:shrinking_of_low_loss_region}. We first generate a single master dataset $\mathcal{D}_{\mathrm{mod1997}}(x^y, 1997)$, and use the same tokenizer for all derived datasets. For each $n \leq 1997$, we then define $\mathcal{D}_{\mathrm{mod1997}}(f,n)$ by retaining only those samples whose operands satisfy $x,y \in \{0,\ldots,n-1\}$. The targets remain $f(x,y)\bmod 1997$ for every $n$; thus, $n$ indexes the operand-domain size rather than the modulus. Consequently, for any $m<n$,  $\mathcal{D}_{\mathrm{mod1997}}(f,m) \subseteq \mathcal{D}_{\mathrm{mod1997}}(f,n)$. Increasing $n$ therefore adds labeled examples and imposes additional constraints. Accordingly, Equation~\ref{eq:shrinking_of_low_loss_region} suggests that the volume
$V\!\left(\mathrm{LLR}\!\left(\mathcal{M}_{\mathrm{grok}}, \mathcal{D}_{\mathrm{mod1997}}(f,n)\right)\right)$ should be non-increasing as $n$ grows.

\subsection{Alignment of Low-Loss Regions}
The grokking setup has two low-loss regions, $\text{LLR}(\mathcal{M}_{grok},\mathcal{D}_{\mathrm{mod1997},\mathrm{train}}(f,n))$ and $\text{LLR}(\mathcal{M}_{grok},\mathcal{D}_{\mathrm{mod1997},\mathrm{val}}(f,n))$. As $n$ increases, both regions shrink. The key question is how the overlap region, $\text{LLR}(\mathcal{M}, \mathcal{D}_{train} + \mathcal{D}_{val})$, and the non-overlap region, $\text{LLR}(\mathcal{M}, \mathcal{D}_{train} + !\mathcal{D}_{val})$, change during this shrinkage.

\begin{figure}[htbp]
    \centering
    \includegraphics[width=0.5\linewidth]{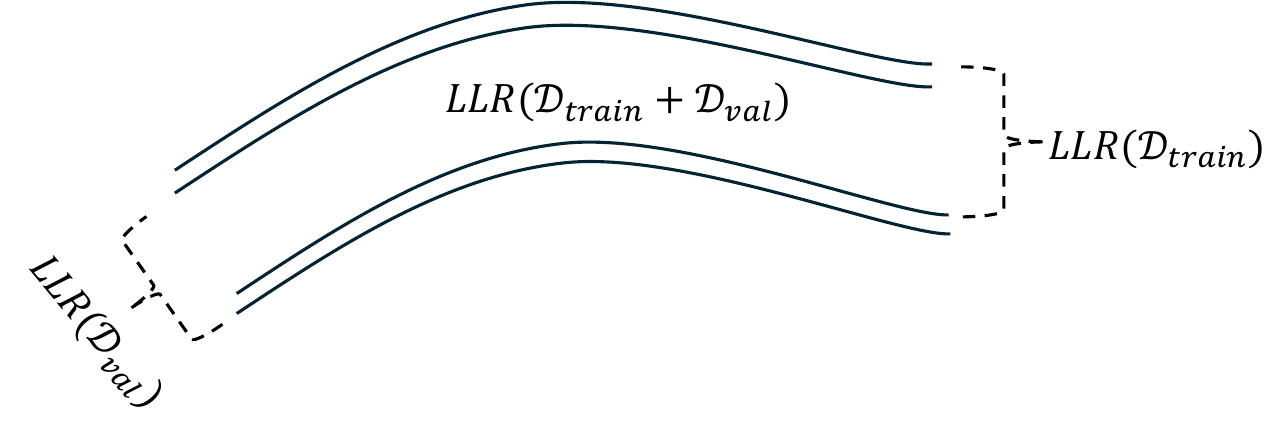}
    \caption{Low-loss regions defined by the training and validation partitions become increasingly aligned as partition size increases. The overlap region is illustrated as a connected path because mode connectivity is empirically verified to persist for $\text{LLR}(\mathcal{M}_{grok}, \mathcal{D}_{train}+\mathcal{D}_{val})$ in Appendix~\ref{appendix:empirical_validation_of_mode_connectivity}.}
    \label{fig:well_aligned_llr}
\end{figure}

Our analysis is that the overlap region should shrink but would never become an empty set as long as the task remains learnable. Otherwise, no model could fit both the training and validation partitions. By contrast, the non-overlap region should decrease as the two low-loss regions shrink. This process is illustrated in Figure~\ref{fig:well_aligned_llr}.

\begin{figure}[htbp]
    \centering
    \includegraphics[width=0.99\linewidth]{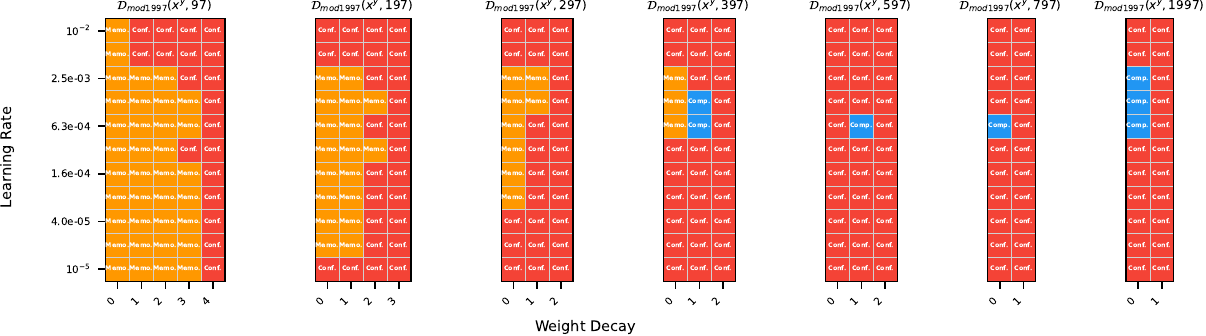}
    \caption{Phase diagrams for $\mathcal{D}_{\mathrm{mod1997}}(x^y,n)$ as the operand-domain size $n$ increases. Each cell represents one learning-rate/weight-decay pair and is classified as \textcolor{ColorComp}{Comprehension}, \textcolor{ColorGrok}{Grokking}, \textcolor{ColorMemo}{Memorization}, or \textcolor{ColorConf}{Confusion} according to the criteria in Appendix~\ref{appendix:learning_phases}. As $n$ increases from $97$ to $1997$, the number of \textcolor{ColorMemo}{Memo.} cells decreases and eventually vanishes, while \textcolor{ColorComp}{Comp.} cells emerge and become more frequent. This shift indicates that fewer training trajectories remain in the training-only low-loss region and more directly reach the overlap of the training and validation low-loss regions, suggesting that the two regions become increasingly aligned. No \textcolor{ColorGrok}{Grokking} cells are observed, consistent with the anti-grokking analysis that the effective information carried by the training partition remains below the threshold required for grokking. We leave the question of how to reproduce \textcolor{ColorGrok}{Grokking} cells in the phase diagram as an after-class exercise. We leave the question of how to reproduce \textcolor{ColorGrok}{Grokking} cells in the phase diagram as an after-class exercise. All the necessary knowledge are provided in this paper, so we encourage readers to work it out before checking the answer in Appendix~\ref{appendix:reproduce_grokking}. Appendix~\ref{appendix:phase_cell_alignment} explains why phase-cell counts provide an indirect probe of alignment.}
    \label{fig:all_phase_diagram}
\end{figure}

To test this analysis, we design an empirical experiment that indirectly probes the overlap region, as shown in Figure~\ref{fig:all_phase_diagram}. The procedure is inspired by prior work studying the effect of training hyperparameters on grokking~\citep{liu2022understandinggrokkingeffectivetheory}. Specifically, we sweep learning rate and weight decay, train on $\mathcal{D}_{\mathrm{mod1997}}(x^y, n)$, and classify the resulting learning phase for each hyperparameter pair\footnote{Our setup is close to Figure 6 of~\citep{liu2022understandinggrokkingeffectivetheory}.}. We then count how many cells belong to each learning phase as $n$ increases. A decrease in \textcolor{ColorMemo}{Memorization} cells together with an increase in \textcolor{ColorComp}{Comprehension} cells suggests that fewer trajectories terminate in the training-only low-loss region and more trajectories directly reach the overlap of $\text{LLR}(\mathcal{D}_{train})$ and $\text{LLR}(\mathcal{D}_{val})$. A complete explanation of why phase-cell counts provide an indirect probe of low-loss-region alignment is given in Appendix~\ref{appendix:phase_cell_alignment}.

We perform these experiments on $\mathcal{D}_{\mathrm{mod1997}}(x^y, n)$ for $n \in \{97,197,297,397,597,797,1997\}$ and the results are shown in Figure~\ref{fig:all_phase_diagram}. The four learning phases follow Figure~\ref{fig:grokking_explaination}, and the empirical classification criteria are given in Appendix~\ref{appendix:learning_phases}. 

This experiment is not a direct measurement of low-loss-region volume. Precisely measuring such a volume is computationally infeasible because it would require exploring a high-dimensional parameter space. The phase diagrams in Figure~\ref{fig:all_phase_diagram} should therefore be interpreted as an approximate probe based on sweeping weight decay and learning rate. \RevisionStart As a complementary diagnostic, Appendix~\ref{appendix:misalignment_measurement} applies a symmetric train/validation measurement to the same nested modular-dataset family to provide an approximate quantification of the misalignment level. \RevisionEnd

Even as approximate probes, the diagrams make the shrinking and alignment trends visible. As $n$ increases, the number of \textcolor{ColorMemo}{Memo.} cells decreases and eventually reaches zero, indicating that fewer training trajectories terminate in the training-only low-loss region. At the same time, \textcolor{ColorComp}{Comp.} cells appear and become more frequent, indicating that a larger fraction of trainable trajectories directly reaches the overlap of $\text{LLR}(\mathcal{M}, \mathcal{D}_{train})$ and $\text{LLR}(\mathcal{M}, \mathcal{D}_{val})$. Together, these trends suggest that the two low-loss regions become increasingly aligned as the nested dataset grows. The prevalence of \textcolor{ColorConf}{Conf.} cells at larger $n$ is consistent with the additional data constraints shrinking the accessible low-loss regions. Appendix~\ref{appendix:effect_of_model_size} shows the complementary effect that larger models make the low-loss regions more misaligned. Appendix~\ref{appendix:phase_diagrams_non_nested_modular_datasets} provides a corresponding comparison using the non-nested datasets $\mathcal{D}_{\mathrm{mod}}(x^y,m)$.

We make several additional experimental arrangements to obtain the results in Figure~\ref{fig:all_phase_diagram}:
\begin{itemize}
    \item The function $f$ affects the geometry of the low-loss regions. We turn to use $f(x,y)=x^y$ because it has a much smaller overlap low-loss region than $f(x,y)=x+y$, see Appendix~\ref{appendix:comparison_of_phase_diagram_x_plus_y_and_x_power_of_y}.
    \item All cells within the same panel are obtained using the same training and validation partitions, generated by the random-partition policy.
    \item We strictly control randomness during training, see Appendix~\ref{appendix:additional_arrangements_in_phase_diagram_experiments} for details.
\end{itemize}

\section{Discussion}
\label{sec:discussion}
% This section discusses the implications and limitations of the preceding analysis and empirical results.

\paragraph{The Cause of Grokking}
Prior work has linked grokking to several factors, including Fourier circuit~\citep{nanda2023progressmeasuresgrokkingmechanistic}, numerical stability~\citep{prieto2025grokkingedgenumericalstability}, weight decay~\citep{xu2026grokgrokkingprovablegrokking} and the Slingshot effect~\citep{thilak2022the}.

Instead, these factors can be separated into primary and secondary factors. The primary factor is the misalignment of low-loss regions, which is determined by the model architecture $\mathcal{M}$ and dataset $\mathcal{D}$, see Equation~\ref{eq:define_LLR_from_model_architecture_dataset}. The secondary factor is the training trajectory, or more precisely, how optimization approaches the overlap of the low-loss regions, see Figure~\ref{fig:grokking_explaination}. This trajectory is jointly controlled by weight decay, learning rate, weight initialization, and related hyperparameters. These factors are secondary because they cannot produce grokking when the primary factor already yields well-aligned low-loss regions.

The primary factors ($\mathcal{M}$ and $\mathcal{D}$) also include several implicit choices. For example, $\mathcal{M}$ should include numerical precision, because model weights in empirical experiments have finite precision rather than ideal real-valued coordinates. For modular arithmetic, $\mathcal{D}$ includes the dataset-generating function $f$, which strongly affects the geometry of low-loss regions. Our experiments show that $\mathcal{D}_{mod}(x^2+y^2,97)$ is more aligned than $\mathcal{D}_{mod}(x+y,97)$, see Appendix~\ref{appendix:phase_diagram_of_additional_datasets_97} for details.

\RevisionStart
The cause of grokking can be summarized in Table~\ref{tab:cause_of_grokking}.

\begin{table}[htbp]
\centering
\caption{Schematic summary of the possible learning phases predicted by anti-grokking experiment and the misalignment of the training and validation low-loss regions. Grokking appears only when the training data are sufficient to make the task learnable while the two low-loss regions remain misaligned.}
\label{tab:cause_of_grokking}
\begin{tabular}{ccc}
\toprule
Is training data sufficient relative to the task? & Misaligned LLR & Aligned LLR \\ \hline
Insufficient & \textcolor{ColorComp}{Comp.} \textcolor{ColorConf}{Conf.} \textcolor{ColorMemo}{Memo.} & \textcolor{ColorComp}{Comp.} \textcolor{ColorConf}{Conf.} \\
Sufficient & \textcolor{ColorComp}{Comp.} \textcolor{ColorConf}{Conf.} \textcolor{ColorGrok}{Grok.} 
\textcolor{ColorMemo}{Memo.} & \textcolor{ColorComp}{Comp.} \textcolor{ColorConf}{Conf.} \\
\bottomrule
\end{tabular}
\end{table}

\RevisionEnd

\RevisionStart

\paragraph{Our Analysis Framework Is Generally Applicable}
Our framework only relies on mode connectivity and the common distinction between $\mathcal{D}_{train}$ and $\mathcal{D}_{val}$. Mode connectivity has been widely observed in CNNs and transformers~\citep{tian2025connectingindependentlytrainedmodes, pmlr-v80-draxler18a}, and separate training and validation partitions are standard practice in machine learning. Therefore, grokking should not be limited to modular arithmetic. Whenever the low-loss regions induced by the training and validation partitions are sufficiently misaligned, there should exist a training trajectory that produces grokking, provided that no additional constraints prevent the optimizer from following such a trajectory.

\paragraph{Our Analysis Framework Provides Explanation for Other Grokking Research}
Existing work shows that grokking can also occur when training MLP/ViT models on MNIST and CIFAR datasets~\citep{humayun2024deepnetworksgrok,shabalin2023re,liu2022understandinggrokkingeffectivetheory}. In the MNIST setting, grokking is induced by sampling a 1k training subset, initializing the model and scaling the weights by 9~\citep{shabalin2023re}, and sweeping weight decay and learning rate. In our framework, the 1k training subset makes the low-loss regions more misaligned by reducing the information carried by the training partition. By contrast, weight scaling, weight decay, and learning rate affect the training trajectory approaching the low-loss regions overlap.

\RevisionEnd

% \paragraph{Properties That Are Not Portable To General Datasets}
% Several aspects of the modular-arithmetic setup may not transfer directly to general datasets.
% \begin{enumerate}
%     \item Modular arithmetic datasets are noise-free, whereas real datasets may contain noise. This can make low-loss regions less clean and may change the observed training dynamics.
%     \item The symmetry-based knowledge concept used in Section~\ref{sec:framework_and_anti_grokking} may not have a direct analogue in general datasets.
%     \item We use a $50\%$ training split, which is uncommon in many practical settings.
% \end{enumerate}

% \paragraph{Example of Small-Scale Emergence}
% The number of \textcolor{ColorComp}{Comp.} cells in Figure~\ref{fig:all_phase_diagram} appears to stabilize from $m=597$ to $m=1997$. This suggests that, after $m=597$, the low-loss region may correspond to a stable representation of the function $f(x,y)=x^y$. This observation may be related to \textit{emergence} in Large Language Models: once the dataset is sufficiently large, the model may acquire the underlying structure of the task~\citep{wei2022emergentabilitieslargelanguage,srivastava2023imitationgamequantifyingextrapolating,schaeffer2023emergentabilitieslargelanguage}.

\paragraph{Interpretation of Grokking}
Prior work often treats grokking as a form of emergent generalization~\citep{nanda2023progressmeasuresgrokkingmechanistic, wei2022emergentabilitieslargelanguage}. This description can suggest that, after sufficiently long training, transformers uncover the underlying knowledge hidden in the data. Our results suggest a different interpretation. Since the root cause of grokking is the misalignment of low-loss regions, delayed generalization indicates that the information carried by the training partition is insufficient relative to the model architecture. In this sense, grokking is not necessarily evidence of a hidden ability of neural networks. Rather, it should be viewed as a warning sign that the training data are insufficient for the model to reach stable generalization without passing through a delayed transition.

Similarly, the observation of memorization also suggests that the training dataset is too small relative to the model architecture. This is because both grokking and memorization can only happen when the training and validation low-loss regions are sufficiently misaligned.

\section{Conclusion}
\label{sec:conclusion}
We argue that grokking is governed by two hierarchical factors. The primary factor is the misalignment of the low-loss regions induced by the training and validation partitions, which is determined by the architecture and dataset. Such misalignment is more likely when the information carried by the training partition is small relative to the expressive capacity of the model. The secondary factor is the training trajectory, which is shaped by weight decay, learning rate, initialization, and related hyperparameters. When the primary factors induce well-aligned low-loss regions, grokking cannot be produced by secondary factors alone.

\section*{AI use statement}

In this work, we used generative AI tools to revise our paper to improve readability, implement methods, and write plotting scripts to visualize results and data.
We have not used generative AI tools for formulating mathematical claims, helping develop theoretical models or conceptual frameworks, providing critical ingredients for proving mathematical claims, assisting in the writing of proofs, proposing or refining hypotheses, designing or providing feedback on research methodology or experiments, cleaning and reformatting datasets, supporting qualitative and thematic data analysis, interpreting results, or assisting in the writing of proofs. 
The rest of the required disclosure tasks are not applicable to this work.

Additionally, we used generative AI tools for sourcing/searching for information and summarizing or analyzing existing literature. We have reviewed all AI-assisted work. We checked LLM-generated research ideas for potential plagiarism through a manual
literature survey. We take responsibility for the final content of this work,
including text, claims or artifacts produced with the aid of generative AI.

\section*{Reproducibility statement}

We provide the source code and simulation configuration files as supplementary material, including the settings used for dataset generation and partitioning, model training, hyperparameter sweeps, phase classification, mode-connectivity experiments, and figure generation.

% \subsubsection*{Acknowledgments}
% Use unnumbered third level headings for the acknowledgments. All
% acknowledgments, including those to funding agencies, go at the end of the paper.

\bibliography{iclr2027_conference}
\bibliographystyle{iclr2027_conference}

\newpage
\appendix
\section{Model Architecture and Training Configuration}
\label{appendix:model_arch_and_train_recipe}
We mostly follow the model architecture and training configuration of OpenAI~\citep{power2022grokkinggeneralizationoverfittingsmall}, with slight modifications. We denote the resulting architecture by $\mathcal{M}_{grok}$. The architecture is specified as follows:

\begin{itemize}
    \item Model type: decoder-only causal transformer.
    \item Number of decoder blocks: $2$.
    \item Model width: $d_{\mathrm{model}} = 128$.
    \item Attention heads per block: $4$.
    \item Per-head key/value dimension: $d_k = d_v = 32$.
    \item Feed-forward hidden dimension: $d_{\mathrm{ff}} = 512$.
    \item Feed-forward nonlinearity: ReLU.
    \item Dropout probability: $0.1$.
    \item Vocabulary size used by the model: $2000$.
    \item Input sequence length for binary modular arithmetic data: $6$ tokens.
    \item Positional encoding: fixed sinusoidal encoding.
    \item Attention mask: causal lower-triangular mask.
    \item Normalization: LayerNorm after the attention residual and after the feed-forward residual.
    \item Output head: bias-free linear projection from $128$ features to $2000$ logits.
\end{itemize}

The training configuration is specified as follows:

\begin{itemize}
    \item Optimizer: AdamW.
    \item Learning rate: $10^{-3}$.
    \item Weight decay: $0.0$.
    \item Batch size: $\text{min}(\text{len}(\mathcal{D}_{mod}), 65536)$
    \item AdamW momentum parameters: $\beta_1 = 0.9$, $\beta_2 = 0.98$.
    \item AdamW numerical stability constant: $\epsilon = 10^{-8}$.
    \item Total training epochs: $150{,}000$ for $D_{mod}(x+y,97)$. For other values of $n$, we decrease the number of epochs and increase the total optimizer update step for more complicated tasks: \[E(n)=\max\!\left\{2000,\,\left\lceil \frac{150000}{S(n)^{1.5}}\right\rceil\right\}.\] where\[S(n)=\left\lceil\frac{\mathrm{TrainingSamples}}{\mathrm{BatchSize}}\right\rceil\].
    \item Warmup epochs: $10$.
    \item Minimum learning rate: $10^{-4}$.
    \item Learning-rate schedule: linear warmup followed by cosine annealing.
    \item Warmup schedule: LinearLR with start factor $10^{-8}$ and end factor $1.0$.
    \item Cosine schedule: CosineAnnealingLR with minimum learning rate $10^{-4}$.
    \item Loss function: cross-entropy on the right-hand side tokens after the equals sign.
\end{itemize}

The batch size and training epoch used in Figure~\ref{fig:all_phase_diagram} are presented in Table~\ref{tab:training_schedule}.

\begin{table}[htbp!]
    \centering
    \caption{Batch sizes and training epochs for Figure~\ref{fig:all_phase_diagram}.}
    \label{tab:training_schedule}
    \begin{tabular}{rcc}
        \toprule
        $n$ & Batch size & Epochs \\
        \midrule
        97   & 9{,}409  & 150{,}000 \\
        197  & 38{,}809 & 150{,}000 \\
        297  & 65{,}536 & 150{,}000 \\
        397  & 65{,}536 & 53{,}034 \\
        497  & 65{,}536 & 53{,}034 \\
        597  & 65{,}536 & 28{,}868 \\
        697  & 65{,}536 & 18{,}750 \\
        797  & 65{,}536 & 13{,}417 \\
        897  & 65{,}536 & 8{,}100 \\
        997  & 65{,}536 & 6{,}630 \\
        1997 & 65{,}536 & 2{,}000 \\
        \bottomrule
    \end{tabular}
\end{table}

\section{Empirical Validation of Mode Connectivity for \texorpdfstring{$\text{LLR}(\mathcal{M}_{grok}, \mathcal{D}_{mod}(x+y, 97))$}{LLR(M, Dmod(x+y,97))}}
\label{appendix:empirical_validation_of_mode_connectivity}

\begin{figure}[h!]
    \centering
    \includegraphics[width=0.99\linewidth]{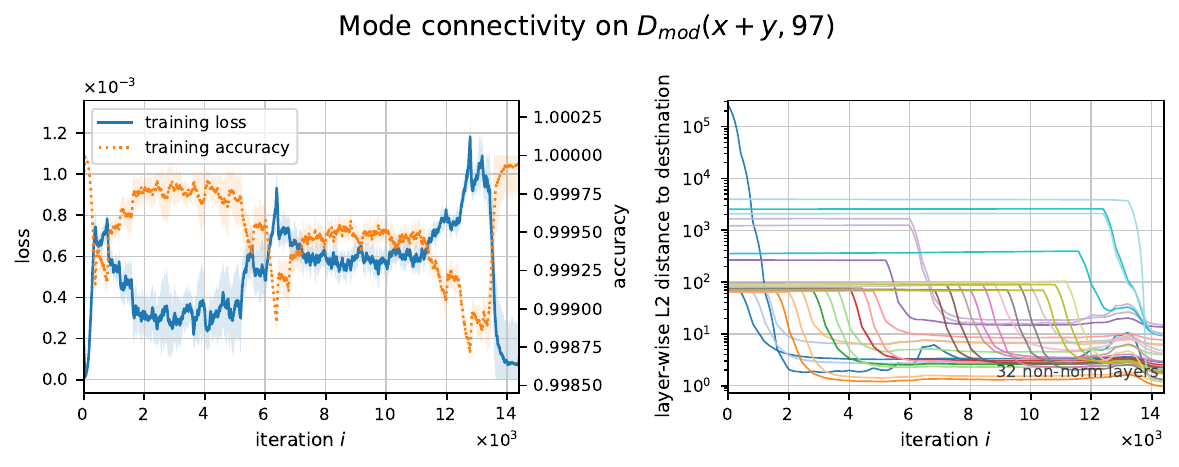}
    \caption{Results of applying LLPF to connect independently trained $\mathcal{M}_{grok}$ models. The left panel reports training accuracy and loss on the full $\mathcal{D}_{mod}(x+y,97)$ dataset along the discovered path. The right panel reports layer-wise $L_2$ distance along the same path. The training loss remains below $1.5\times10^{-3}$, and the $L_2$ distance decreases gradually, supporting mode connectivity for $\mathcal{M}_{grok}$ on $\mathcal{D}_{mod}(x+y,97)$. The final point has nearly the same training accuracy and loss as the starting point because both endpoints are obtained through SGD. The experiment is repeated ten times, and the shaded region in the top panel indicates the standard deviation across runs.}
    \label{fig:addition_97_m2m_10_mode_connectivity}
\end{figure}

We empirically validate mode connectivity for $\mathcal{M}_{grok}$ on $\mathcal{D}_{mod}(x+y, 97)$. For each trial, we train a pair of independently initialized models and then connect the resulting modes using the Low-Loss Path Finding Algorithm (LLPF) introduced in~\citep{tian2025connectingindependentlytrainedmodes}. Figure~\ref{fig:addition_97_m2m_10_mode_connectivity} shows the resulting paths.

The LLPF hyperparameters used to reproduce Figure~\ref{fig:addition_97_m2m_10_mode_connectivity} are listed below. The notation follows Algorithm 1 and Table 4 of~\citep{tian2025connectingindependentlytrainedmodes}.
\begin{itemize}
    \item Phase iterations: $T=400$.
    \item Step parameter: $\mathrm{step}_a=0.002$.
    \item Step parameter: $\mathrm{step}_f=0.002$.
    \item Maximum training rounds: $r=400$.
    \item Loss threshold: $0.001$.
    \item Batch size: $\mathcal{B}=9409$ (full batch).
    \item Optimizer: AdamW with $\eta=0.0001$ and $\beta=(0.9,0.999)$.
\end{itemize}
LLPF updates selected layers in a prescribed order. We therefore move one layer per phase and use the following hierarchical layer order. Layer names follow PyTorch naming conventions:
\begin{enumerate}
    \item Top-level order: $\text{embedding} \rightarrow \text{block0} \rightarrow \text{block1} \rightarrow \text{linear}$.
    
    \item Within each block (block0 and block1): $\text{head0} \rightarrow \text{head1} \rightarrow \text{head2} \rightarrow \text{head3} \rightarrow \text{ffn}$.
    
    \item Within each attention head (head0 to head3): $\text{query} \rightarrow \text{key} \rightarrow \text{value}$.
\end{enumerate}

\RevisionStart

Because each partition is a subset of the full dataset, Equation~\ref{eq:shrinking_of_low_loss_region} suggests that the low-loss region for the full dataset is contained in the low-loss region of either subset, $\mathcal{D}_{mod\_train}(x+y,97)$ or $\mathcal{D}_{mod\_val}(x+y,97)$. Therefore, under the same loss threshold, any low-loss path found for the full dataset is expected to remain low-loss when evaluated on either partition.

\begin{figure}[htbp]
    \centering
    \includegraphics[width=0.99\linewidth]{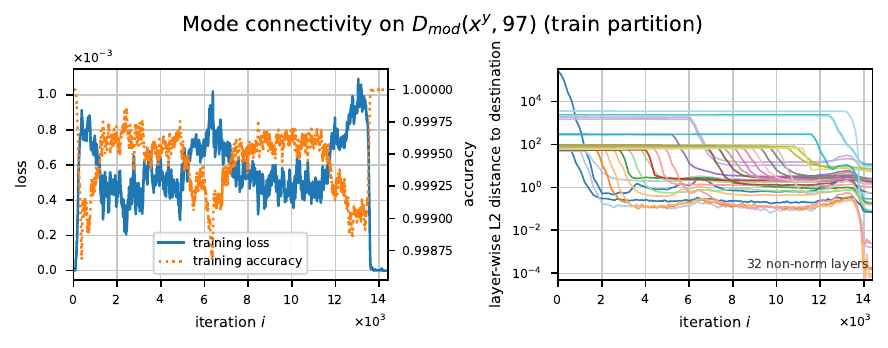}
    \caption{Results of applying LLPF to connect two independently trained $\mathcal{M}_{grok}$ models on the training partition of $\mathcal{D}_{mod}(x^y,97)$. }
    \label{fig:exp_97_m2m_10_mode_connectivity}
\end{figure}

Because Figure~\ref{fig:all_phase_diagram} is based on the $x^y$ task, we also provide an explicit LLPF result for the training partition of $\mathcal{D}_{mod}(x^y,97)$ as a consistency check. This additional experiment directly supports the use of the mode-connectivity interpretation for the partition-specific phase diagrams, see Figure~\ref{fig:exp_97_m2m_10_mode_connectivity}.

\RevisionEnd

\section{Phase Cells as an Indirect Probe of Alignment}
\label{appendix:phase_cell_alignment}

The phase diagram provides an indirect way to probe the alignment of the training and validation low-loss regions. Each cell corresponds to one pair of training hyperparameters, such as the learning rate and weight decay, and therefore to one optimization trajectory under the fixed model, dataset, initialization, and training protocol. The phase label summarizes the qualitative behavior of that trajectory according to the criteria in Appendix~\ref{appendix:learning_phases}.

\begin{figure}[htbp]
    \centering
    \includegraphics[width=0.8\linewidth]{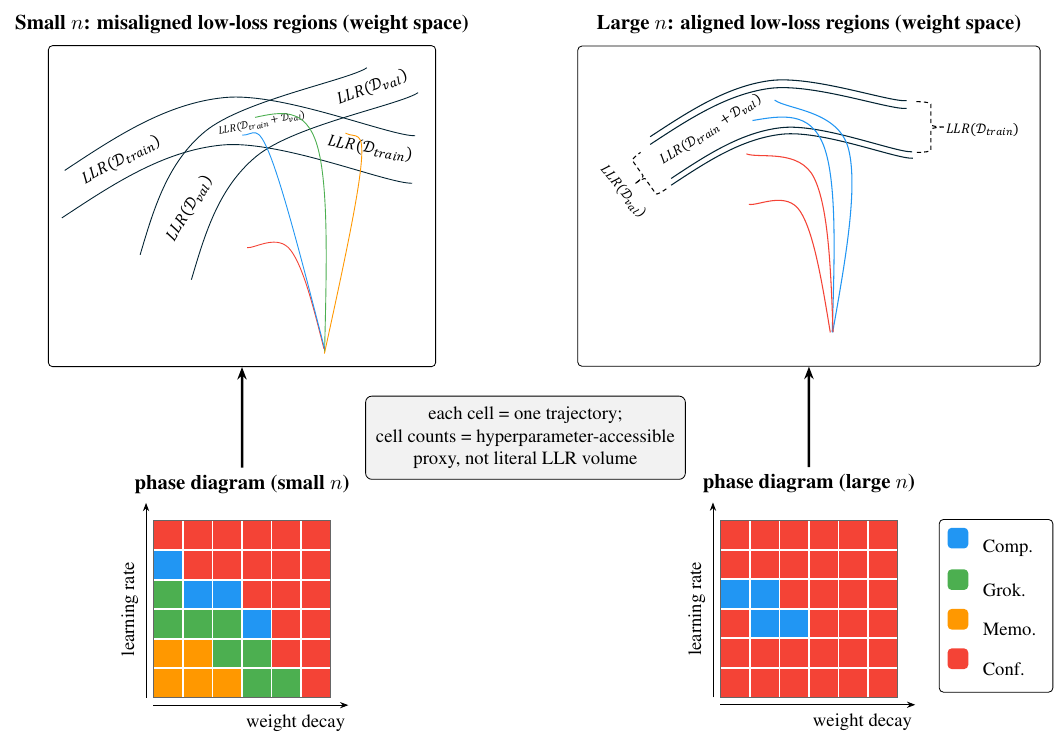}
    \caption{Schematic explanation of why phase-cell counts can serve as an indirect probe of low-loss-region alignment. The upper panels show a misaligned case, in which the training and validation low-loss regions have a relatively large training-only portion, and an aligned case, in which their overlap is relatively larger. Each cell in the corresponding phase diagram represents one learning-rate/weight-decay pair and therefore one training trajectory. As the nested dataset grows, the low-loss regions contract while the optimization problem remains otherwise controlled. The resulting changes in the phase-cell distribution provide a hyperparameter-accessible proxy for alignment rather than a literal measurement of low-loss-region volume.}
    \label{fig:phase_cell_alignment}
\end{figure}

The upper panels of Figure~\ref{fig:phase_cell_alignment} schematically show two possible geometries. When the training and validation low-loss regions are misaligned, the training-only region is relatively large and different hyperparameter choices can produce \textcolor{ColorMemo}{Memorization} or \textcolor{ColorGrok}{Grokking}. When the two regions become more aligned, the training-only region becomes smaller relative to their overlap, so fewer trajectories are expected to produce these two phases. The corresponding phase diagrams in the lower panels therefore contain fewer \textcolor{ColorMemo}{Memorization} and \textcolor{ColorGrok}{Grokking} cells and a larger fraction of trajectories classified as \textcolor{ColorComp}{Comprehension} or \textcolor{ColorConf}{Confusion}.

In our construction, the full labeled datasets are nested: for $m<n$, $\mathcal{D}_{\mathrm{mod1997}}(f,m)\subseteq\mathcal{D}_{\mathrm{mod1997}}(f,n)$. Therefore, increasing $n$ adds constraints while retaining all constraints from the smaller dataset, and the corresponding low-loss region is expected to contract rather than move to an unrelated part of parameter space. This inclusion relation is a geometric heuristic when the loss is averaged over samples, as discussed in Section~\ref{sec:alignment_of_low_loss_region}, but it provides the basis for comparing the phase diagrams across $n$.

Moreover, with the model architecture, initialization, optimizer, randomness control, and training budget fixed, SGD remains driven, on average, toward lower training loss. Changing $n$ therefore changes the set of constraints and the target low-loss region, but does not replace the optimization problem with an unrelated one. We consequently expect the training trajectories to change primarily through their access to the shrinking low-loss regions, rather than through arbitrary changes in trajectory geometry. This controlled change is what makes the distribution of phase cells useful for comparing alignment across dataset sizes.

\section{Empirical Criteria for Learning Phases}
\label{appendix:learning_phases}

Table~\ref{tab:learning-phases} defines the empirical criteria used to assign each training trajectory to a learning phase.

\begin{table}[htbp]
\centering
\caption{Empirical definitions of the four learning phases.}
\label{tab:learning-phases}
\setlength{\tabcolsep}{2pt}
\begin{tabular}{lccc}
\toprule
& \multicolumn{3}{c}{Criteria} \\
\cmidrule(lr){2-4}
Phase
& \shortstack{Final train.\\acc. $\geq 95\%$}
& \shortstack{Final val.\\acc. $\geq 95\%$}
& \shortstack{Val. acc. at\\$t_{\mathrm{train}}^{95\%} \geq 50\%$} \\
\midrule
\textcolor{ColorComp}{Comprehension} & Yes & Yes & Yes \\
\textcolor{ColorGrok}{Grokking}      & Yes & Yes & No  \\
\textcolor{ColorMemo}{Memorization}  & Yes & No  & Not applicable \\
\textcolor{ColorConf}{Confusion}     & No  & No  & Not applicable \\
\bottomrule
\end{tabular}
\end{table}

\section{Phase Diagrams for Non-Nested Modular Datasets}
\label{appendix:phase_diagrams_non_nested_modular_datasets}

We also report phase diagrams for the conventional, non-nested datasets $\mathcal{D}_{\mathrm{mod}}(x^y,m)$ with $m\in\{97,197,297,397,597,797,1997\}$. Unlike the fixed-modulus family in the main experiment, changing $m$ changes the operand domain and labels, so these panels provide a qualitative cross-dataset comparison rather than a nested-dataset test. We use the same phase-classification criteria as in the main experiment.

\begin{figure}[htbp]
    \centering
    \includegraphics[width=0.99\linewidth]{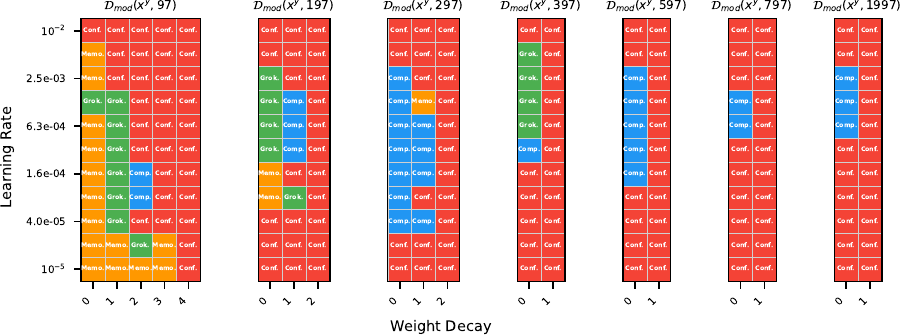}
    \caption{Phase diagrams for the conventional modular datasets $\mathcal{D}_{\mathrm{mod}}(x^y,m)$ as the modulus $m$ increases. Each cell represents one learning-rate/weight-decay pair and is classified using the criteria in Appendix~\ref{appendix:learning_phases}. As $m$ increases, \textcolor{ColorMemo}{Memorization} cells decrease and eventually disappear, while \textcolor{ColorComp}{Comprehension} cells emerge and become more prominent; \textcolor{ColorGrok}{Grokking} cells are mainly observed at intermediate values of $m$. The resulting shift toward \textcolor{ColorComp}{Comprehension} and \textcolor{ColorConf}{Confusion} is qualitatively consistent with increasing low-loss-region alignment. Because the datasets are not nested across $m$, however, this figure should be interpreted as a cross-dataset comparison rather than direct evidence of monotonic low-loss-region shrinkage.}
    \label{fig:all_phase_diagram_old}
\end{figure}

\RevisionStart

\section{Reproduce Grokking in Phase Diagram}
\label{appendix:reproduce_grokking}

Figure~\ref{fig:all_phase_diagram} shows no \textcolor{ColorGrok}{Grokking} cells. Here is the after-class exercise: how can we recover \textcolor{ColorGrok}{Grokking} cells while keeping the phase-diagram procedure fixed?

The answer requires separating task learnability from low-loss-region alignment. Recall our discussion of effective training information entropy: the training partition needs to carry sufficient task-relevant information for the model to recover the full task. Let $\mathcal{I}_{\mathrm{train}}$ denote the task-relevant information carried by the training partition, $C_{\mathrm{task}}$ denote the complexity of the target task, and $C_{\mathrm{model}}$ denote the effective capacity of the model. These quantities are qualitative bookkeeping variables rather than formal measurements in this paper. A task must be sufficiently learnable, meaning that $\mathcal{I}_{\mathrm{train}}$ is large enough relative to $C_{\mathrm{task}}$ ($\frac{\mathcal{I}_{\mathrm{train}}}{C_{\mathrm{task}}}$ needs to be larger than a threshold). At the same time, grokking requires the training partition to remain insufficient relative to the model capacity, so that the training and validation low-loss regions remain misaligned ($\frac{\mathcal{I}_{\mathrm{train}}}{C_{\mathrm{model}}}$ needs to be smaller than a threshold).

However, when $\mathcal{I}_{\mathrm{train}}$ increases, it is possible to reach the alignment threshold $\frac{\mathcal{I}_{\mathrm{train}}}{C_{\mathrm{model}}}$ first and reach the task-learnability threshold $\frac{\mathcal{I}_{\mathrm{train}}}{C_{\mathrm{task}}}$ later. In this case, grokking will never happen because the low-loss regions are already well-aligned when the training partition finally carries sufficient information to recover the task, just like Figure~\ref{fig:all_phase_diagram}.

Thus, the answer to reproducing grokking is to lower the complexity of the target task. This increases $\frac{\mathcal{I}_{\mathrm{train}}}{C_{\mathrm{task}}}$ and allows the task-learnability threshold to be reached before the alignment threshold $\frac{\mathcal{I}_{\mathrm{train}}}{C_{\mathrm{model}}}$.

\begin{figure}[htbp]
    \centering
    \includegraphics[width=0.99\linewidth]{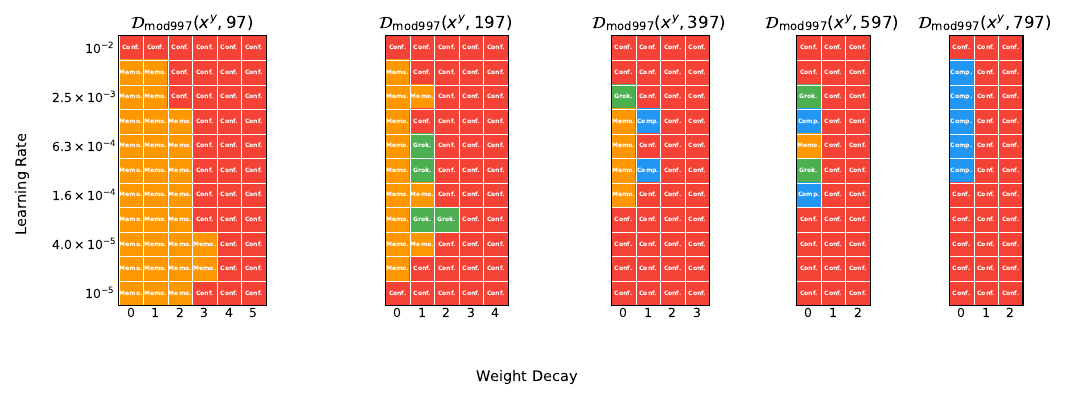}
    \caption{Phase diagrams for dataset $\mathcal{D}_{\mathrm{mod997}}(x^y,n)$, where $n$ is the operand-domain size and all targets are computed modulo $997$. Each cell is classified using the criteria in Appendix~\ref{appendix:learning_phases}. \textcolor{ColorGrok}{Grokking} cells reappear for the intermediate sizes $n=197$, $397$, and $597$ under the same model and phase-diagram procedure used in Figure~\ref{fig:all_phase_diagram}.}
    \label{fig:grokking_is_back}
\end{figure}

We demonstrate this solution by changing the fixed modulus of the nested dataset from $\mathcal{D}_{\mathrm{mod1997}}(x^y,n)$ to $\mathcal{D}_{\mathrm{mod997}}(x^y,n)$. We repeat the same learning-rate/weight-decay sweep for $n\in\{97,197,397,597,797\}$. As shown in Figure~\ref{fig:grokking_is_back}, \textcolor{ColorGrok}{Grokking} cells reappear for the intermediate operand-domain sizes $n=197$, $397$, and $597$.

\RevisionEnd

\section{Experimental Arrangements for Phase Diagram Sweeps}
\label{appendix:additional_arrangements_in_phase_diagram_experiments}

The phase-diagram experiment in Figure~\ref{fig:all_phase_diagram} uses several additional arrangements compared with the grokking experiment in Figure~\ref{fig:grokking_setup}(b):
\begin{itemize}
    \item The CosineAnnealingLR scheduler is replaced with a constant learning-rate schedule to isolate the effect of the swept hyperparameters.
    \item The batch size is set to $\text{min}(\text{len}(\mathcal{D}), 65536)$ to limit GPU memory consumption.
    \item All cells for the same dataset start from the same initialized weights to reduce randomness across the sweep.
    \item The random seed is fixed across cells for the same dataset, reducing stochasticity from mini-batching and SGD.
    \item For each learning rate, we iterate over weight decay values $\lambda=0 \dots 9$. If two consecutive weight-decay values yield \textcolor{ColorConf}{confusion}, the remaining cells for that learning rate are also treated as \textcolor{ColorConf}{confusion}. This shortcut is motivated by empirical work showing that loss often follows a convex relationship with weight decay~\citep{lewkowycz2021trainingdynamicsdeepnetworks,dangelo2024needweightdecaymodern}.
    \item We use the tokenizer file from the largest modulus. For every operand-domain size $n$, we use the tokenizer copied from the fixed-modulus-$1997$ master dataset $\mathcal{D}_{\mathrm{mod1997}}(x^y,1997)$.
\end{itemize}

\section{Phase-Diagram Comparison Between \texorpdfstring{$\mathcal{D}_{mod}(x+y,997)$}{Dmod(x+y,997)} and \texorpdfstring{$\mathcal{D}_{mod}(x^y,997)$}{Dmod(x_power_y,997)}}
\label{appendix:comparison_of_phase_diagram_x_plus_y_and_x_power_of_y}

\begin{figure}[htbp]
    \centering
    \includegraphics[width=0.3\linewidth]{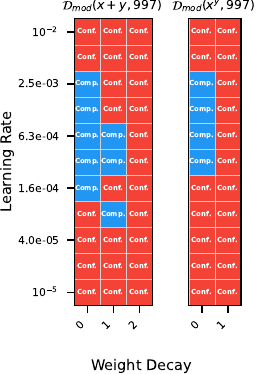}
    \caption{Comparison of phase diagrams for $\mathcal{D}_{mod}(x+y,997)$ and $\mathcal{D}_{mod}(x^y,997)$. The visible trainable region is larger for $\mathcal{D}_{mod}(x+y,997)$, indicating a larger low-loss region.}
    \label{fig:appendix_llr_comparision}
\end{figure}

Figure~\ref{fig:appendix_llr_comparision} illustrates that the low-loss region of $\mathcal{D}_{mod}(x^y,997)$ is smaller than that of $\mathcal{D}_{mod}(x+y,997)$. Both datasets are randomly partitioned into training and validation sets using a $50\%$ split.

\section{Anti-Grokking Ablation by Same-Partition Ratio}
\label{appendix:anti_grokking_example_with_other_ratios}

\begin{figure}[ht]
    \centering
    \includegraphics[width=0.99\linewidth]{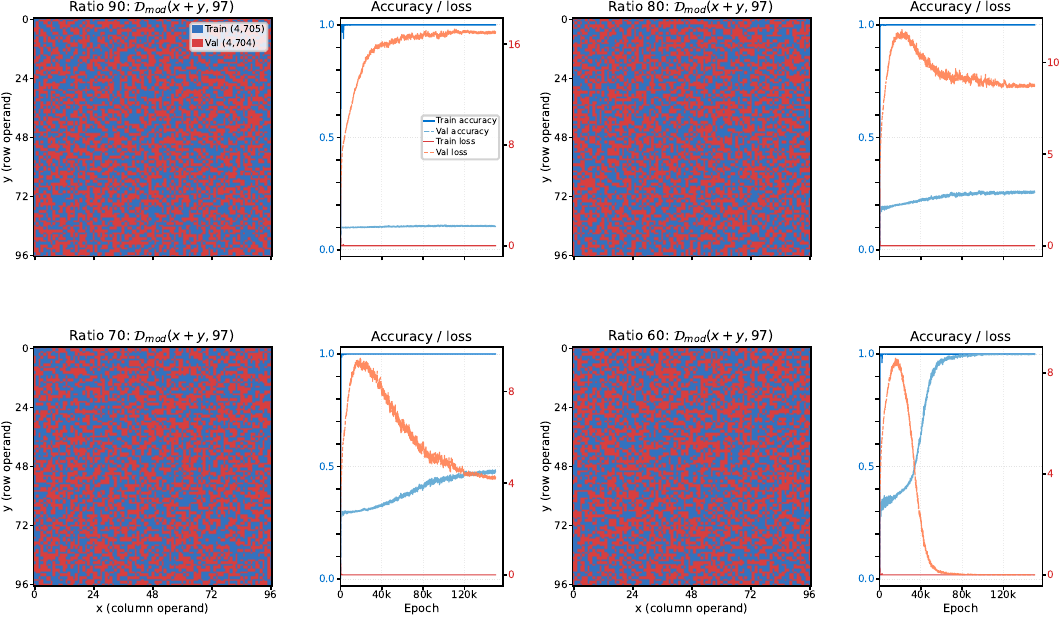}
    \caption{Grokking gradually recovers as the same-partition ratio decreases from $100\%$ toward lower ratios.}
    \label{fig:anti_grokking_other_ratio}
\end{figure}

Figure~\ref{fig:anti_grokking_other_ratio} shows that grokking can recover when the same-partition ratio is reduced from the anti-grokking value of $100\%$. At $100\%$, every paired sample located at $(a,b)$ and $(b,a)$ belongs to the same partition. Lowering this ratio moves the split away from the fully symmetry-preserving anti-grokking construction.

\subsection{Symmetry-aware training coverage}
\label{appendix:effective_information_entropy}
The main text argues that the raw size of the training partition is not the fundamental quantity governing grokking. Instead, the relevant quantity is the effective information entropy carried by that partition. To make this idea operational, we use the symmetry of the modular-addition task, $f(a,b)=f(b,a)$. Let $\mathcal{T}$ denote the set of operand pairs in the training partition, and let $\mathcal{O}(a,b)=\{(a,b),(b,a)\}$ denote the corresponding symmetry orbit. We call an orbit \emph{covered} when $\mathcal{O}(a,b)\cap\mathcal{T}\neq\varnothing$; thus, observing one member of the orbit provides information about both ordered samples. The symmetry-aware training coverage is then the fraction of symmetry orbits covered by $\mathcal{T}$:

\begin{align}
    C_{\mathrm{sym}}(\mathcal{T})
    =
    \frac{\left|\left\{\mathcal{O}(a,b):\mathcal{O}(a,b)\cap\mathcal{T}\neq\varnothing\right\}\right|}
    {\left|\left\{\mathcal{O}(a,b)\right\}\right|}.
\end{align}

\begin{figure}[ht]
    \centering
    \includegraphics[width=0.8\linewidth]{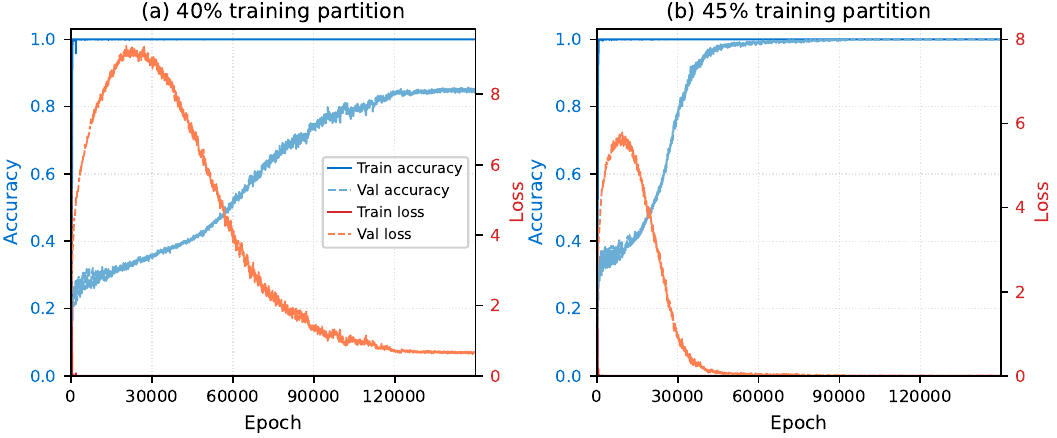}
    \caption{Training dynamics for the normal grokking case on $\mathcal{D}_{\mathrm{mod}}(x+y,97)$. With a $40\%$ training partition (a), validation accuracy remains below one and no complete grokking transition is observed. With a $45\%$ training partition (b), training accuracy reaches one first and validation accuracy rises substantially later, producing a clear grokking trajectory.}
    \label{fig:training_coverage}
\end{figure}

We compare this coverage with the raw training-partition size in two experiments. First, under the random-partition policy, we vary the training-partition size. As shown in Figure~\ref{fig:training_coverage}, $40\%$ training does not produce a complete grokking trajectory, whereas $45\%$ training produces a clear delayed rise in validation accuracy. The corresponding coverages are $0.637$ and $0.695$, respectively. Second, in the anti-grokking experiment, we vary the fraction of symmetric pairs assigned to the same partition, as shown in Figure~\ref{fig:anti_grokking_other_ratio}. No complete grokking trajectory is observed at a same-partition ratio of $70\%$, whereas a clear grokking trajectory is recovered at $60\%$; the corresponding coverages are $0.648$ and $0.680$. Thus, the transition intervals in the two experiments overlap substantially. In particular, the coverage at $45\%$ training in the random-partition experiment is close to the coverage at a $60\%$ same-partition ratio in the anti-grokking experiment, even though the raw partition parameters differ. This agreement suggests that symmetry-aware training coverage captures task-relevant information more accurately than the raw number of training samples and can serve as an empirical proxy for effective information entropy. This interpretation is empirical and limited to the modular addition task. A formal entropy measure that incorporates task knowledge remains an interesting direction for future work.

\begin{table}[ht!]
    \centering
    \caption{Symmetry-aware training coverage for the random-partition grokking experiments and the same-partition-ratio anti-grokking experiments. For the random partitions, the reported values are the expected coverages $p/m+(1-1/m)(2p-p^2)$ with $m=97$ and training fraction $p$; for the symmetry-controlled partitions, coverage is computed directly from the realized split using the same symmetry-closure definition. In the random-partition setting, the transition from no complete grokking to grokking occurs between $40\%$ and $45\%$ training, corresponding to coverages of $0.637$ and $0.695$. In the anti-grokking setting, the corresponding transition occurs between same-partition ratios of $70\%$ and $60\%$, corresponding to coverages of $0.648$ and $0.680$. The close coverage values suggest that symmetry-aware coverage is a more informative comparison across the two experimental controls than the raw partition parameters.}
    \label{tab:symmetry_aware_training_coverage}
    \begin{tabular}{cc|cc}
        \toprule
        \multicolumn{2}{c|}{Grokking case} &
        \multicolumn{2}{c}{Anti-grokking case} \\
        Train partition (\%) &
        Training coverage &
        Same-partition ratio (\%) &
        Training coverage \\
        \midrule
        50 & 0.747 & 100 & 0.500 \\
        \textbf{45} & \textbf{0.695} & 90 & 0.550 \\
        40 & 0.637 & 80 & 0.599 \\
        35 & 0.575 & 70 & 0.648 \\
        30 & 0.508 & \textbf{60} & \textbf{0.680} \\
        25 & 0.436 & 50 & 0.747 \\
        \bottomrule
    \end{tabular}
\end{table}

\clearpage

\RevisionStart

\section{Why Optimization Moves from \texorpdfstring{$\text{LLR}(\mathcal{D}_{train})$}{LLR(D\_train)} to \texorpdfstring{$\text{LLR}(\mathcal{D}_{train}+\mathcal{D}_{val})$}{LLR(D\_train+D\_val)}}
\label{appendix:cause_of_moving_to_overlap}

The geometric picture in Section~\ref{sec:framework_and_anti_grokking} separates grokking into two parts: reaching the low-loss region defined by the training partition and then reaching its overlap with the validation low-loss region. Entering the first region does not terminate optimization. A low-loss region is a sublevel set of the training objective, not a set of parameter vectors with identical loss, and training accuracy records only whether the predicted class is correct. Consequently, $100\%$ training accuracy does not imply that the training cross-entropy is zero or that its gradient is zero.

\begin{figure}[htbp]
    \centering
    \includegraphics[width=0.9\linewidth]{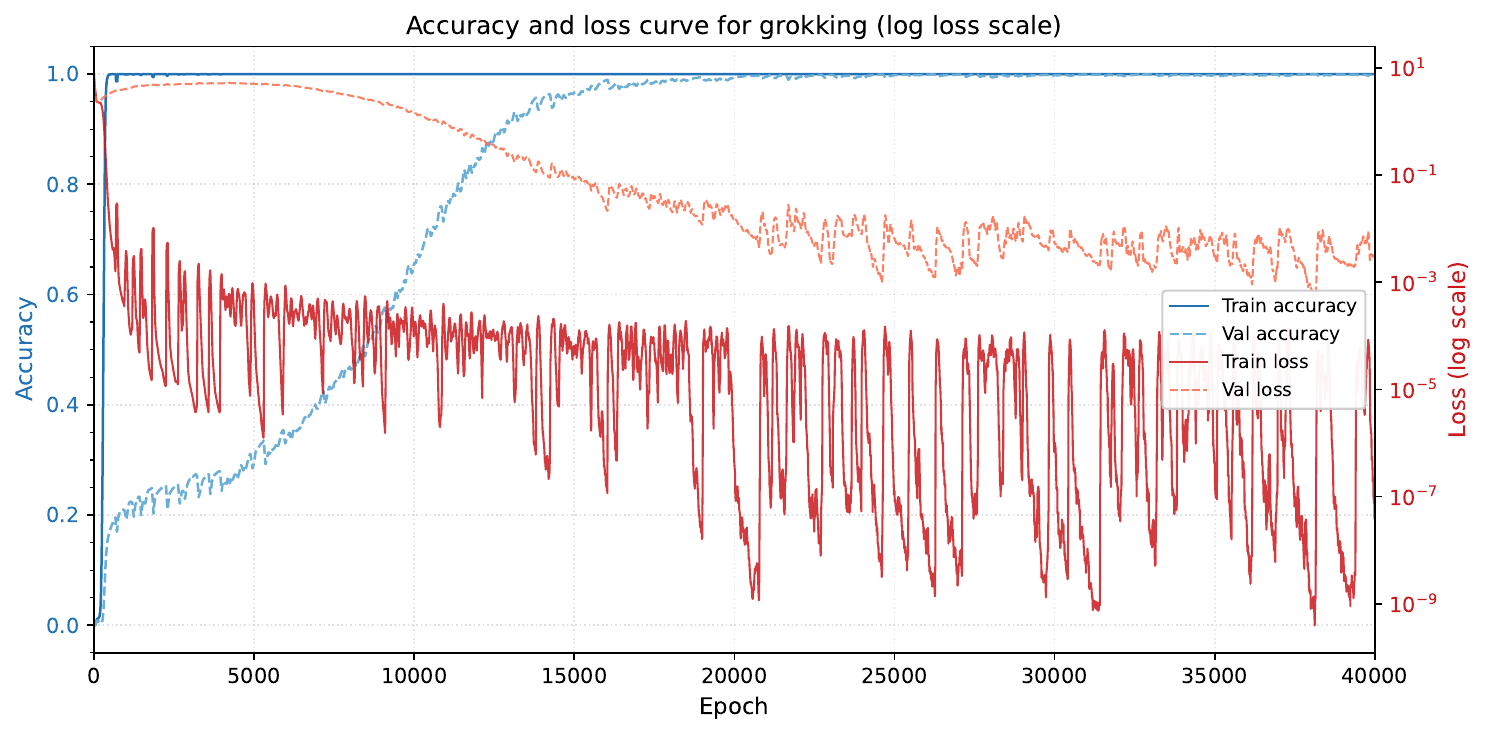}
    \caption{Log-scale loss view of the grokking run in Figure~\ref{fig:grokking_setup}(b). Training accuracy reaches $100\%$ around epoch $1000$, while the training cross-entropy continues to decrease from approximately $0.002$ toward $0.0002$ as validation accuracy approaches $0.99$. The loss curves contain optimization oscillations, so the decrease is most clearly seen in their envelope.}
    \label{fig:grokking_log_loss}
\end{figure}

Figure~\ref{fig:grokking_log_loss} provides evidence for this mechanism in the baseline grokking run. Training accuracy reaches one early, but the training loss continues to decrease on a log scale, from approximately $2\times10^{-3}$ to $2\times10^{-4}$ by the time validation accuracy approaches $0.99$.

This behavior may be less visible on noisy real-world datasets. Label noise can impose a nonzero irreducible loss floor, so a loss near $0.002$ may already be close to the best achievable value and optimization may eventually plateau. The modular arithmetic dataset is noise-free, so it does not impose such a statistical floor; the optimizer can continue reducing the training loss to much smaller values. This distinction helps explain why the post-saturation decrease is especially clear in the present experiment.

\RevisionEnd

\RevisionStart

\section{Measurement of Low-Loss-Region Misalignment}
\label{appendix:misalignment_measurement}

The phase-cell analysis in Section~\ref{sec:alignment_of_low_loss_region} provides an indirect probe of low-loss-region alignment. We additionally use an optimization-based procedure to measure how easily a model can reach a region with low training loss but high validation loss to indicate the volume of the region.

For a loss threshold $\mathcal{L}_{th}$, the volume of the train-only region can be written as
\begin{align}
    V_{\mathrm{train-only}}(\mathcal{L}_{th})
    =
    \int_{\mathbb{R}^{d}}
    \mathbf{1}\!\left[
        \mathcal{L}_{\mathrm{train}}(\theta)\leq\mathcal{L}_{th}
        \ \land\
        \mathcal{L}_{\mathrm{val}}(\theta)>\mathcal{L}_{th}
    \right]
    \,d\theta .
\end{align}
Computing this quantity would require integrating over the high-dimensional parameter space and is computationally infeasible. We therefore use a local optimization proxy. Starting from the same initial model, we deliberately decrease the training loss while increasing the validation loss. The ease with which this procedure finds a train-only solution provides evidence about the accessible part of
$\text{LLR}(\mathcal{D}_{train}+!\mathcal{D}_{val})$.

For each nested dataset
$\mathcal{D}_{\mathrm{mod1997}}(x^y,n)$, the examples are shuffled once with a fixed split seed and divided into two equal partitions, $A$ and $B$. We run both orientations: $A$ as the training partition and $B$ as the validation partition, followed by $B$ as the training partition and $A$ as the validation partition. Both orientations start from the same initial model parameters.

During each paired mini-batch update, the procedure computes the training gradient $g_X$ and validation gradient $g_Y$. For each parameter tensor, the two gradients are normalized independently. The normalized gradients are then combined as
\begin{align}
    d_j
    =
    0.6\,\widehat{g}_{X,j}
    -
    0.4\,\widehat{g}_{Y,j},
\end{align}
where $j$ indexes the parameter tensors. The combined direction is rescaled to the norm of the original training gradient before the optimizer update. Because AdamW subtracts the supplied update direction, this procedure decreases the training loss while increasing the validation loss. The full algorithm is available in Algorithm~\ref{alg:misalignment_measurement}.

\begin{algorithm}[htbp]
    \caption{Optimization-based measurement of train-only accessibility}
    
    \small
    \label{alg:misalignment_measurement}
    \begin{algorithmic}[1]
        \REQUIRE Dataset $\mathcal{D}_n$, split seed $s$, initial model $\theta_0$, epochs $T$
        \STATE Shuffle $\mathcal{D}_n$ using seed $s$ and split it into equal partitions $A$ and $B$.
        \FOR{$(X,Y)\in\{(A,B),(B,A)\}$}
            \STATE Restore $\theta\leftarrow\theta_0$.
            \STATE Treat $X$ as training data and $Y$ as validation data.
            \FOR{$t=1,\ldots,T$}
                \FOR{each paired mini-batch $(b_X,b_Y)$}
                    \STATE Compute $g_X=\nabla_{\theta}\mathcal{L}_{X}(b_X)$ and $g_Y=\nabla_{\theta}\mathcal{L}_{Y}(b_Y)$.
                    \STATE Normalize $g_{X,j}$ and $g_{Y,j}$ independently for every parameter tensor $j$.
                    \STATE Form $d_j=0.6\,\widehat{g}_{X,j}-0.4\,\widehat{g}_{Y,j}$.
                    \STATE Rescale $d_j$ to $\lVert g_{X,j}\rVert$.
                    \STATE Apply one AdamW update using the direction $d$.
                \ENDFOR
            \ENDFOR
            \STATE Evaluate final loss and accuracy on $X$ and $Y$.
        \ENDFOR
        \RETURN Final training and validation losses and accuracies for both orientations.
    \end{algorithmic}
\end{algorithm}

\begin{figure}[ht]
    \centering
    \includegraphics[width=0.9\linewidth]{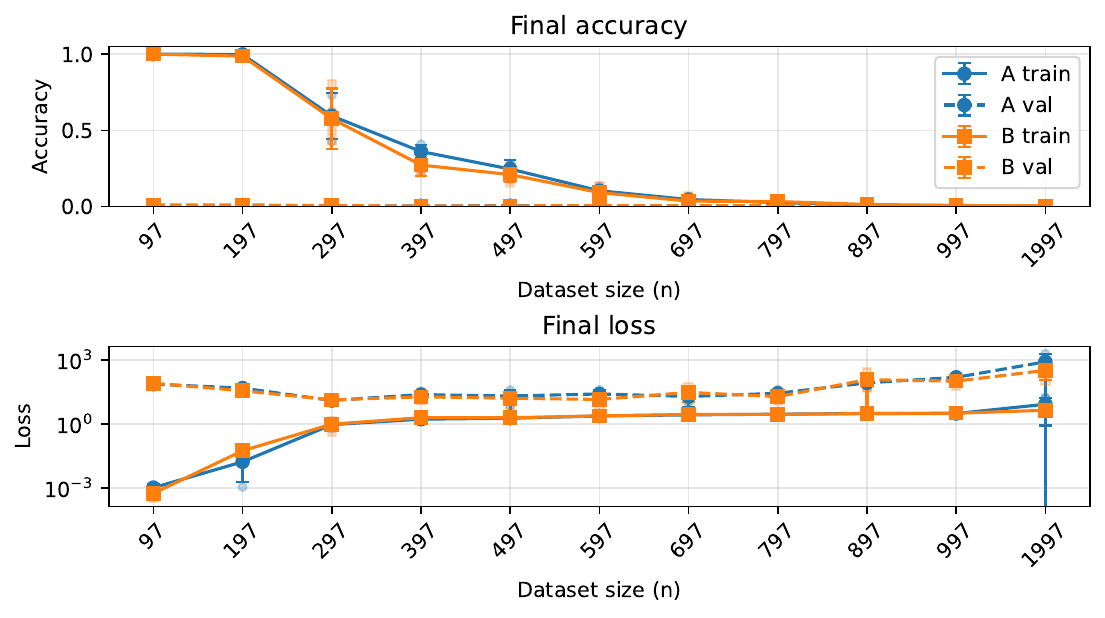}
    \caption{Final training and validation accuracies (top panel) and loss (bottom panel) from the symmetric misalignment measurement on $\mathcal{D}_{\mathrm{mod1997}}(x^y,n)$. The x-axis (dataset size) denotes the value $n$. Each point is the mean over three repeated runs, error bars show one sample standard deviation, and faint markers show individual repeats. Solid curves show training accuracy for the two train/validation orientations. Dashed curves show the corresponding validation accuracies. A larger train–validation accuracy gap indicates greater misalignment between the two low-loss regions.}
    \label{fig:misalignment_mod1997}
\end{figure}

We repeat the measurement three times for every $n\in\{97,197,\dots997,1997\}$ and report the mean and sample standard deviation. Figure~\ref{fig:misalignment_mod1997} shows the final accuracies produced by these runs. At $n=97$, training accuracy is approximately one while validation accuracy remains near zero, indicating that the optimization procedure can readily find a train-only solution. As $n$ increases, both training and validation accuracies decrease and become nearly zero by $n=1997$. Under the fixed model and optimization budget, train-only solutions therefore become increasingly difficult to find. This trend is consistent with a shrinking accessible non-overlap region or increasing alignment, but it may also reflect a general loss of trainability. The result should therefore be interpreted as an optimization-based proxy rather than a direct measurement of low-loss-region volume.

\RevisionEnd

\section{Additional Phase Diagrams for \texorpdfstring{$m=997$}{m=997}}
\label{appendix:phase_diagram_of_additional_datasets_997}

This section reports phase diagrams for four additional datasets: $\mathcal{D}_{mod}(x+y,997)$, $\mathcal{D}_{mod}(xy,997)$, $\mathcal{D}_{mod}(x^2+2y^2,997)$, and $\mathcal{D}_{mod}(x^2+2y^2+xy,997)$. These diagrams use a narrower weight-decay range than Figure~\ref{fig:all_phase_diagram}, giving higher resolution and revealing more \textcolor{ColorComp}{Comp.} cells.

\begin{figure}[htbp]
    \centering
    \includegraphics[width=0.6\linewidth]{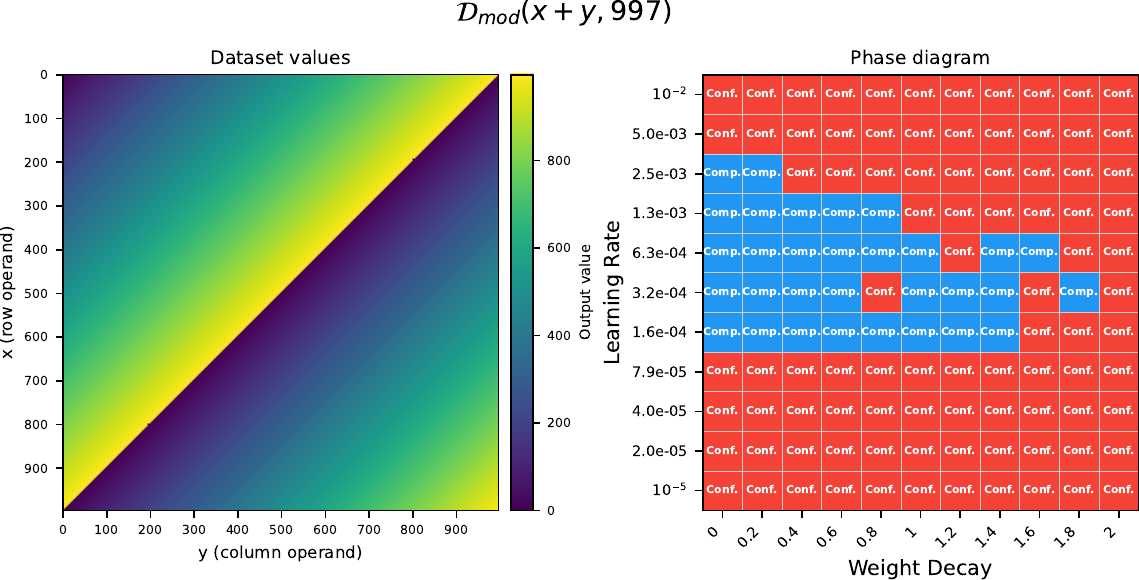}
    \caption{Dataset illustration (left) and phase diagram (right) for $\mathcal{D}_{mod}(x+y,997)$.}
    \label{fig:phase_diagram_x_plus_y}
\end{figure}
\begin{figure}[htbp]
    \centering
    \includegraphics[width=0.6\linewidth]{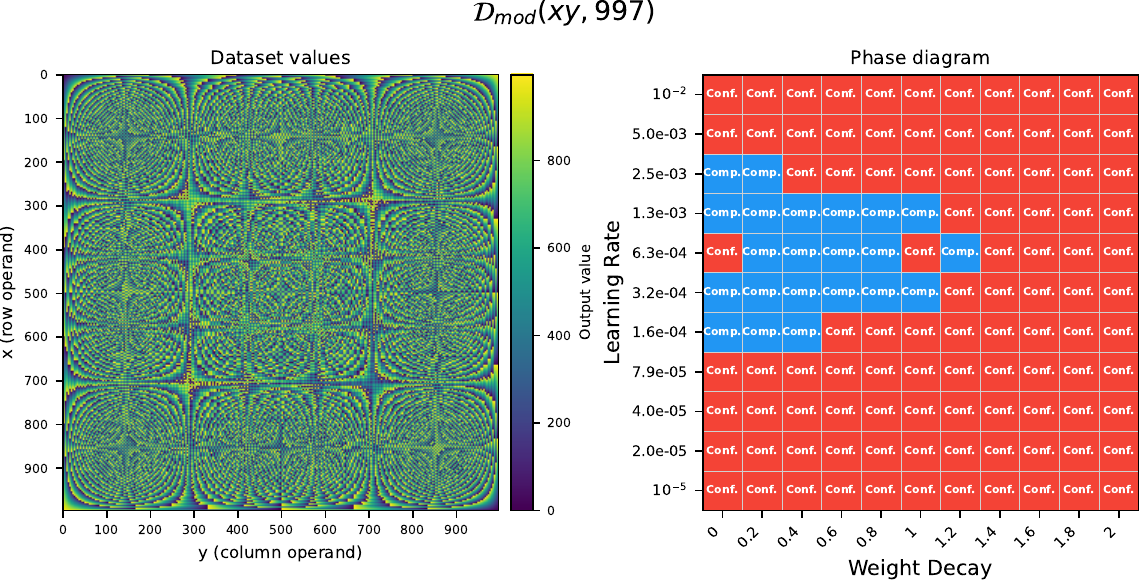}
    \caption{Dataset illustration (left) and phase diagram (right) for $\mathcal{D}_{mod}(xy,997)$.}
    \label{fig:phase_diagram_xy}
\end{figure}

\begin{figure}[htbp]
    \centering
    \includegraphics[width=0.6\linewidth]{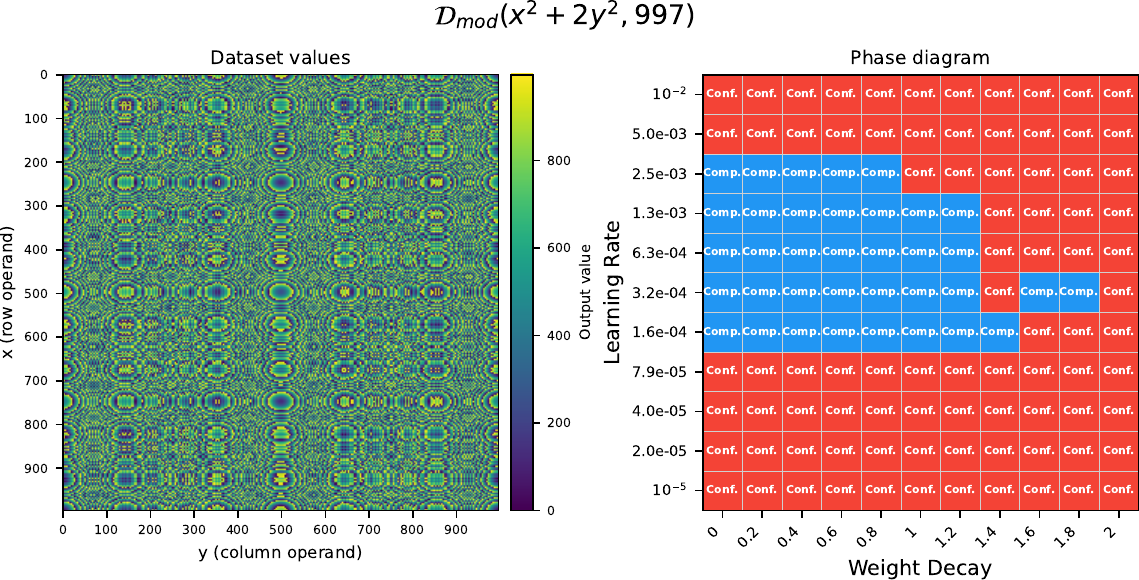}
    \caption{Dataset illustration (left) and phase diagram (right) for $\mathcal{D}_{mod}(x^2+2y^2,997)$.}
    \label{fig:phase_diagram_x2_plus_2y2}
\end{figure}
\begin{figure}[htbp]
    \centering
    \includegraphics[width=0.6\linewidth]{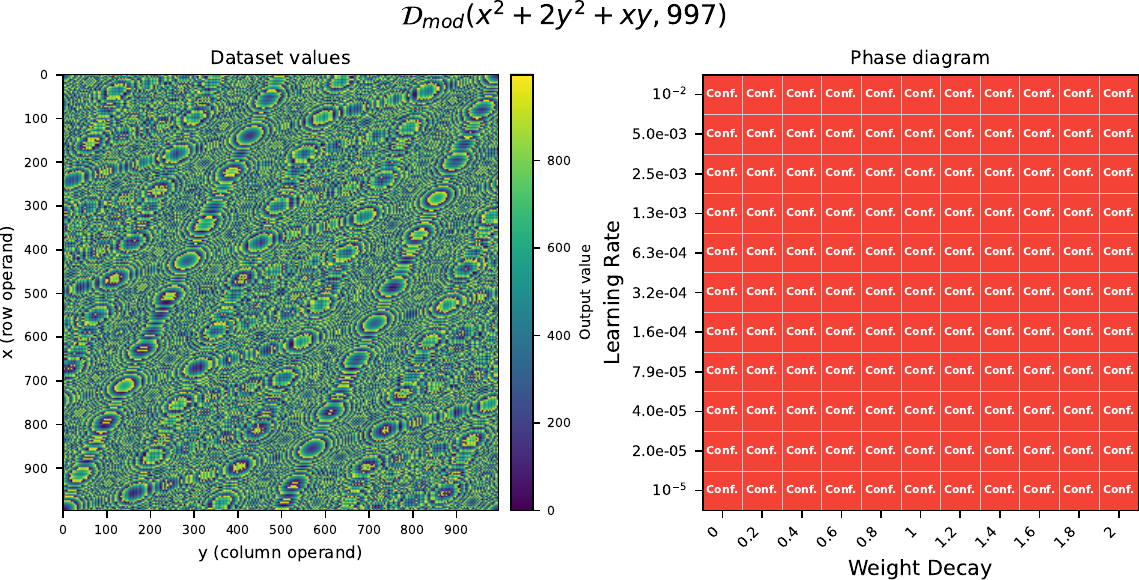}
    \caption{Dataset illustration (left) and phase diagram (right) for $\mathcal{D}_{mod}(x^2+2y^2+xy,997)$.}
    \label{fig:phase_diagram_x2_plus_2y2_plus_xy}
\end{figure}

Figures~\ref{fig:phase_diagram_x_plus_y}--\ref{fig:phase_diagram_x2_plus_2y2_plus_xy} show that combining learnable functions does not guarantee trainability. At $m=997$, both $f_1(x,y)=xy$ and $f_2(x,y)=x^2+2y^2$ are trainable because they produce a visible number of \textcolor{ColorComp}{Comp.} cells. However, their sum $f(x,y)=f_1+f_2=x^2+2y^2+xy$ is not trainable under the same sweep, because no \textcolor{ColorComp}{Comp.} cells are observed.

\section{Additional Phase Diagrams for \texorpdfstring{$m=97$}{m=97}}
\label{appendix:phase_diagram_of_additional_datasets_97}

This section reports phase diagrams for $\mathcal{D}_{mod}(x+y,97)$, $\mathcal{D}_{mod}(x^2+y^2,97)$, and $\mathcal{D}_{mod}(2x^2+y,97)$. Under $m=97$, the low-loss regions for $f(x,y)=x^2+y^2$ are the most aligned among these three functions, those for $f(x,y)=x+y$ are intermediate, and those for $f(x,y)=2x^2+y$ are the least aligned.

\begin{figure}[htbp]
    \centering
    \includegraphics[width=0.6\linewidth]{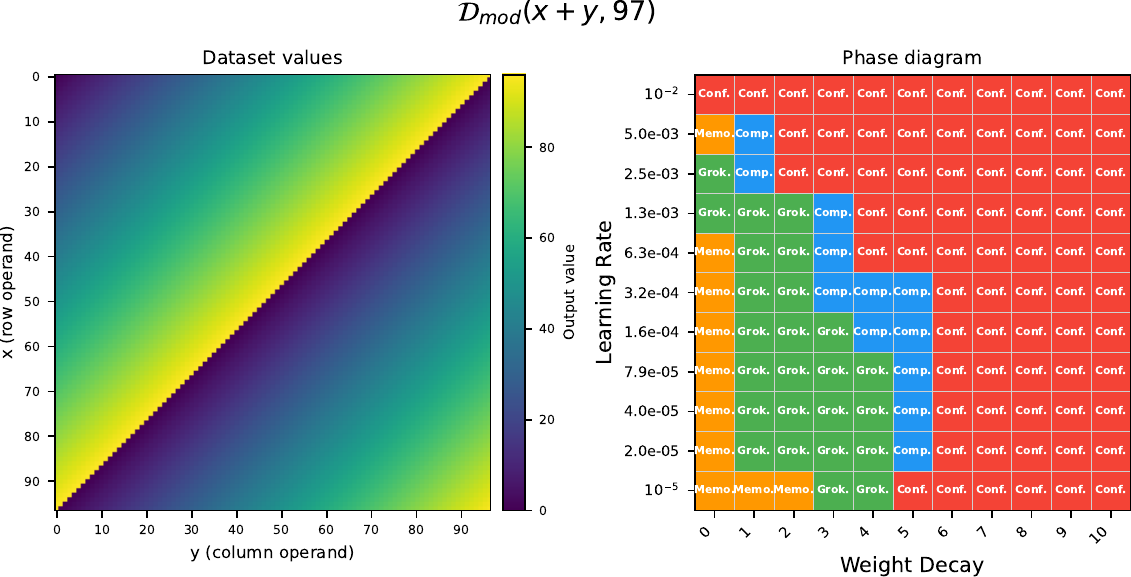}
    \caption{Dataset illustration (left) and phase diagram (right) for $\mathcal{D}_{mod}(x+y,97)$.}
    \label{fig:phase_diagram_two_panel_x_plus_y}
\end{figure}
\begin{figure}[htbp]
    \centering
    \includegraphics[width=0.6\linewidth]{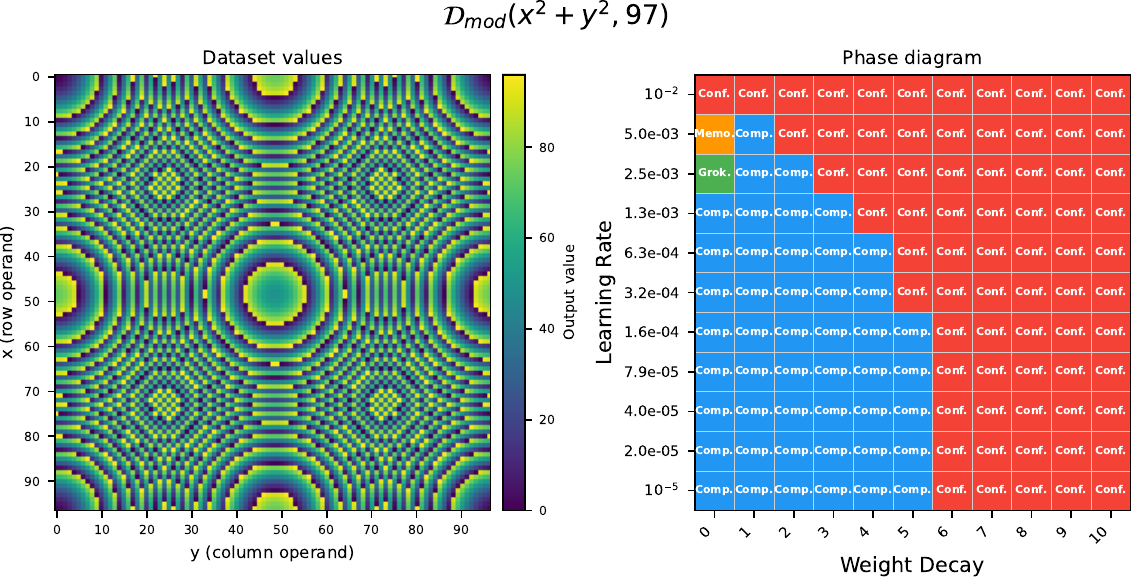}
    \caption{Dataset illustration (left) and phase diagram (right) for $\mathcal{D}_{mod}(x^2+y^2,97)$.}
    \label{fig:phase_diagram_two_panel_x2_plus_y2}
\end{figure}
\begin{figure}[htbp]
    \centering
    \includegraphics[width=0.6\linewidth]{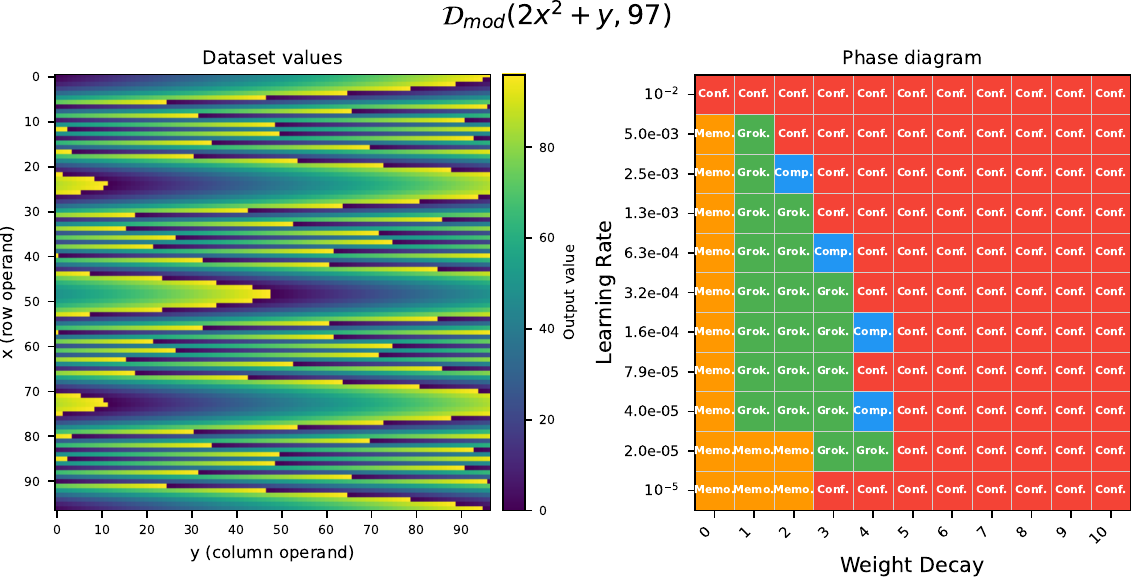}
    \caption{Dataset illustration (left) and phase diagram (right) for $\mathcal{D}_{mod}(2x^2+y,97)$.}
    \label{fig:phase_diagram_two_panel_2x2_plus_y}
\end{figure}

\newpage
\section{Effect of Model Size on Phase Diagrams}
\label{appendix:effect_of_model_size}
This section illustrates how model size influences low-loss-region alignment.

\begin{figure}[htbp]
    \centering
    \includegraphics[width=0.55\linewidth]{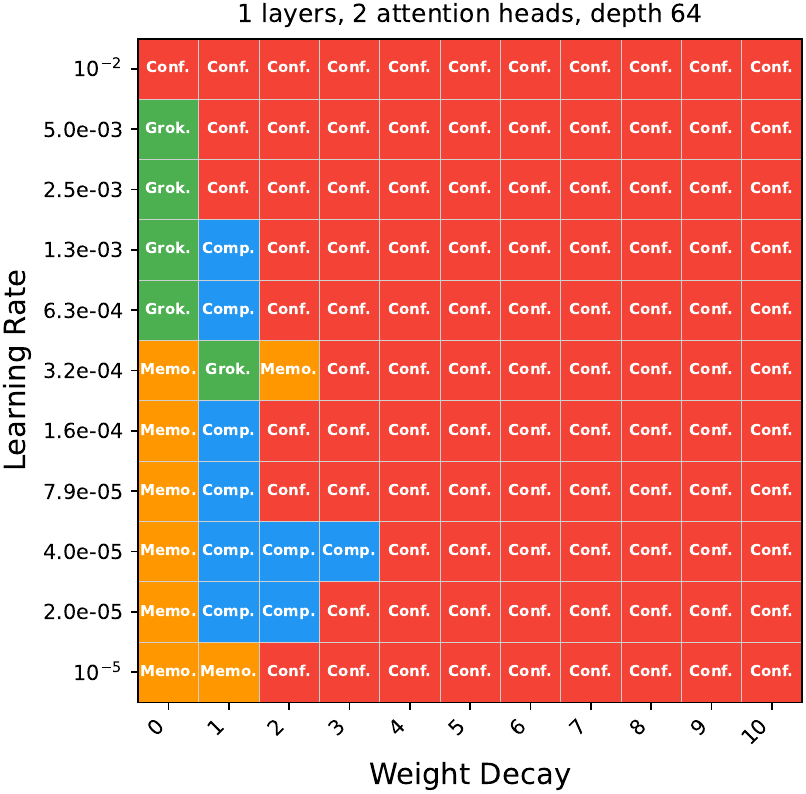}
    \caption{Smaller model with $1$ decoder layer, $2$ attention heads, and width $64$. The fraction of \textcolor{ColorComp}{Comprehension} cells is higher than in the baseline model, indicating that the low-loss regions are more aligned.}
    \label{fig:phase_diagram_small_model}
\end{figure}
\begin{figure}[htbp]
    \centering
    \includegraphics[width=0.55\linewidth]{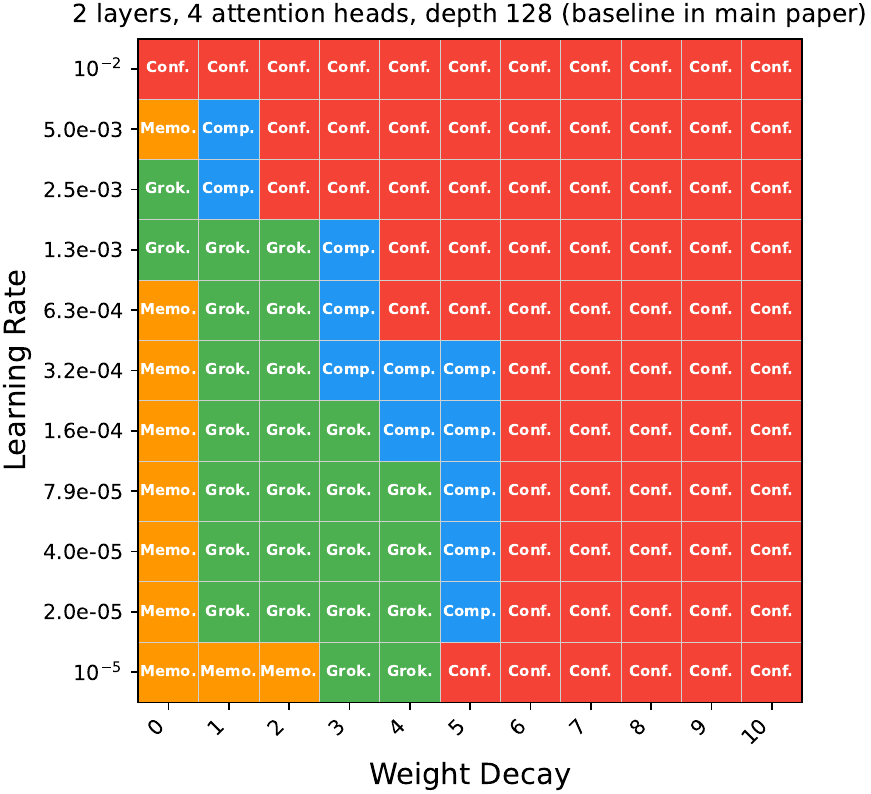}
    \caption{Baseline model with $2$ decoder layers, $4$ attention heads per layer, and width $128$.}
    \label{fig:phase_diagram_baseline_model}
\end{figure}
\begin{figure}[htbp]
    \centering
    \includegraphics[width=0.55\linewidth]{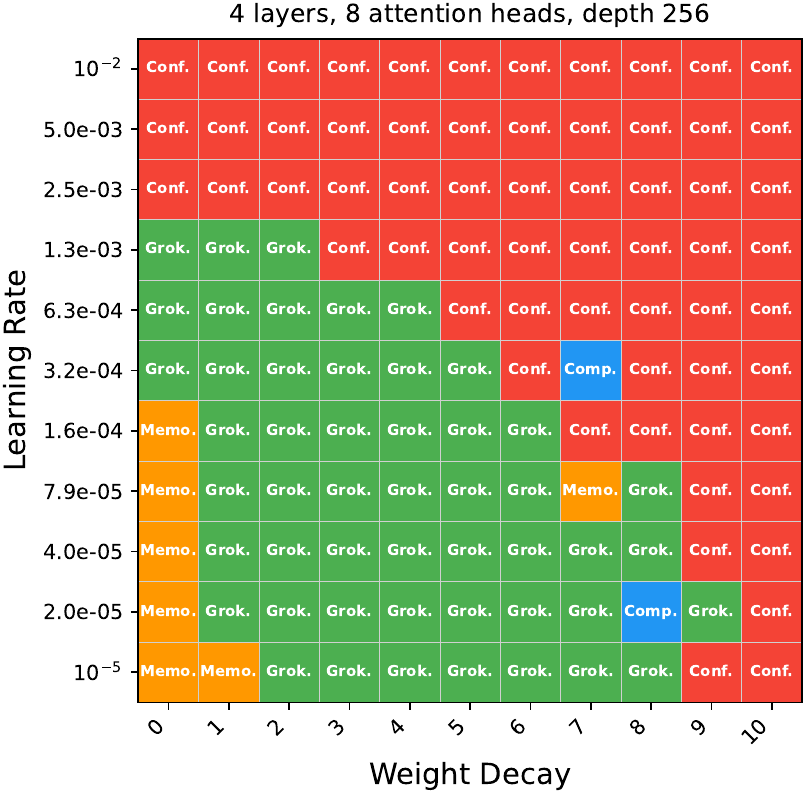}
    \caption{Larger model with $4$ decoder layers, $8$ attention heads per layer, and width $256$. \textcolor{ColorComp}{Comprehension} cells are rare, indicating that the low-loss regions are more misaligned than in the baseline model.}
    \label{fig:phase_diagram_large_model}
\end{figure}

The baseline $\mathcal{M}_{grok}$ model has $2$ decoder layers, $4$ attention heads per layer, and width $128$. We construct smaller and larger variants by multiplying these values by $0.5$ and $2$, respectively. We then train each model on $\mathcal{D}_{mod}(x+y,97)$ to examine low-loss-region alignment. Since a larger model has greater representational capacity, increasing model size can play a role similar to reducing the effective dataset size, making the low-loss regions more misaligned. The results in Figures~\ref{fig:phase_diagram_small_model}--\ref{fig:phase_diagram_large_model} support this analysis.

\section{Singular Learning Theory}
\label{appendix:slt}
Singular Learning Theory(SLT)~\citep{9812468} treats layered neural networks as singular models, implying that directions can exist in $\mathbb{R}^d$ along which a model moves while preserving loss. This view suggests that SGD solutions do not simply lie at the bottom of isolated basins, as in traditional loss-landscape visualizations~\citep{Visualizing_the_Loss_Landscape_of_Neural_Nets}, but instead lie along extended valleys~\citep{carroll2021phase}. SLT provides theoretical motivation for studying low-loss-region geometry, but it does not by itself establish global path connectivity.

Mode Connectivity (MC) provides complementary empirical evidence by searching for low-loss paths between independently trained solutions~\citep{pmlr-v80-draxler18a, tian2025connectingindependentlytrainedmodes}. We use LLPF for this purpose. Repeated LLPF success supports Conjecture~\ref{conj:low_loss_region_is_path_connected}, which states that the relevant full-dataset low-loss region is path-connected, but it does not prove the conjecture. Thus, SLT provides theoretical motivation for the valley-like geometry, while MC provides an empirical test of the connectivity assumption.

\section{Mode Connectivity vs Linear Mode Connectivity}
\label{appendix:mc_vs_lmc}
We distinguish Mode Connectivity (MC) from Linear Mode Connectivity (LMC). LMC works do not investigate mode connectivity for independently trained models~\citep{zhou2023goinglinearmodeconnectivity}. LMC typically studies whether two low-loss models can be connected by a linear low-loss path when the models are obtained by \textbf{spawning} or \textbf{permutation}. \textbf{Model spawning} initializes and trains one model for a few epochs, copies it into two descendants, and then trains the descendants independently to convergence~\citep{pmlr-v119-frankle20a, fort2020deeplearningversuskernel}. \textbf{Model permutation} trains two models independently and then permutes the neurons of one model to align it with the other, producing a functionally equivalent representation~\citep{ainsworth2023git, entezari2022the}.

\section{Limitations}
\label{appendix:limitations}
Several aspects of the modular-arithmetic setup may not transfer directly to general datasets.
\begin{enumerate}
    \item Modular arithmetic datasets are noise-free, whereas real datasets may contain noise. This can make low-loss regions less clean and may change the observed training dynamics.
    \item The symmetry-based knowledge concept used in Section~\ref{sec:framework_and_anti_grokking} may not have a direct analogue in general datasets.
    \item We use a $50\%$ training split, which is uncommon in many practical settings.
\end{enumerate}

\section{Computational Resources}

We report the GPU cost of producing the main results. A single grokking experiment, corresponding to one cell in Figure~\ref{fig:all_phase_diagram}, takes approximately $25$ minutes on an NVIDIA RTX 5090 GPU for $150{,}000$ epochs. Sweeping a full $11\times11=121$ grid for one dataset takes approximately $2.1$ days. To reduce this cost, we skip the remaining cells for a learning rate after two consecutive weight-decay values fall in the \textcolor{ColorConf}{confusion} region. With this optimization, the runtime for one phase diagram is reduced to approximately one day.

\end{document}